\documentclass[a4paper,fleqn]{cas-dc}

\usepackage[numbers]{natbib}

\usepackage{graphicx}
\usepackage{amsmath}

\usepackage{amssymb}
\usepackage{booktabs}
\usepackage{multirow}
\usepackage{ulem}
\usepackage{makecell}
\usepackage{float}
\usepackage{subfigure}

\def\tsc#1{\csdef{#1}{\textsc{\lowercase{#1}}\xspace}}
\tsc{WGM}
\tsc{QE}
\begin{document}
\let\WriteBookmarks\relax
\def\floatpagepagefraction{1}
\def\textpagefraction{.001}

\shorttitle{An Experimental Study of Semantic Continuity for Deep Learning Models}

\shortauthors{Shangxi Wu \emph{et al.}}  

\title [mode = title]{An Experimental Study of Semantic Continuity for Deep Learning Models}  

% \tnotemark[<tnote number>] 



\author[1]{Shangxi Wu}[orcid=0000-0002-6826-6396]
% % Corresponding author indication
% \cormark[5]
% % Footnote of the first author
% % Email id of the first author
\ead{wushangxi@bjtu.edu.cn}
% % URL of the first author
% \ead[url]{<URL>}
% % Credit authorship
% % eg: \credit{Conceptualization of this study, Methodology, Software}
% \credit{<Credit authorship details>}

\author[2]{Dongyuan Lu}
\ead{ludy@uibe.edu.cn}
\fnmark[*]

\author[1]{Xian Zhao}
\ead{lavender.zxshane@gmail.com}

\author[1]{Lizhang Chen}
\ead{lzchen@utexas.edu}

\author[1]{Jitao Sang}
\ead{jtsang@bjtu.edu.cn}

% % Address/affiliation
\affiliation[1]{organization={Beijing Key Lab of Traffic Data Analysis and Mining},
            addressline={Beijing Jiaotong University}, 
            city={Beijing},
            postcode={100091}, 
            country={China}}

\affiliation[2]{organization={School of Information Technology and Management},
            addressline={University of International Business and Economics},
            city={Beijing},
            postcode={100029}, 
            country={China}}

% % Corresponding author text
\cortext[4]{*Corresponding author: Dongyuan Lu}

\begin{abstract}
Deep learning models often face the issue of semantic discontinuity, where minor perturbations in the input space lead to significant semantic-level interference in the model's output. We argue that this semantic discontinuity stems from inappropriate training targets and contributes to well-known problems such as adversarial robustness and interpretability. To address this, we first perform data analysis to provide evidence of semantic discontinuity in existing deep learning models. We then propose a simple semantic continuity constraint that theoretically enables models to achieve smooth gradients and learn features oriented towards semantics. Both qualitative and quantitative experiments demonstrate that models incorporating semantic continuity successfully reduce reliance on non-semantic information, thereby improving adversarial robustness, interpretability, model transferability, and reducing data bias.
\end{abstract}


\maketitle

% Main text
\section{Introduction}

\begin{figure}[!t]
   \begin{center}
      \includegraphics[width=0.95\linewidth]{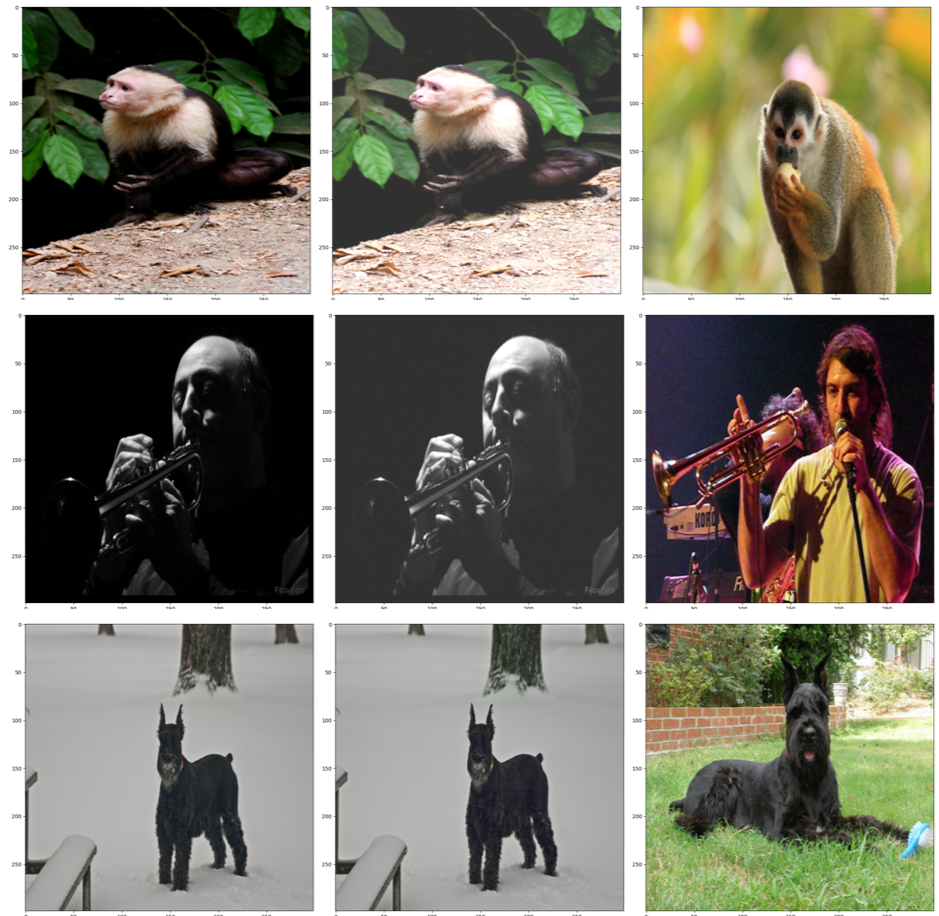}
   \end{center}
      \caption{
        The first column shows the original sample $x_{A}$. The second column displays the data-augmented sample $x_{A'}$, which has been slightly adjusted for brightness, contrast, saturation, and hue. The third column presents the image $x_{B}$ from the same category. Notably, in the commonly used ResNet-50 pre-trained model, the representation distance between $x_{A}$ and $x_{A'}$ is greater than that between $x_{A}$ and $x_{B}$.  
         }
   \label{AAB}
\end{figure}

Deep learning models can achieve state-of-the-art performance across a wide range of computer vision tasks. From supervised learning and unsupervised learning to the now popular self-supervised learning, new training paradigms have progressively improved the efficiency of utilizing training data. However, the existence of issues such as adversarial examples makes us realize that the current training paradigms still do not make sufficient use of datasets. Adversarial images, which appear nearly identical to the original images, can cause significant changes in model output. In this paper, we find that many common non-semantic perturbations can also lead to semantic-level interference in model outputs, as illustrated in Fig.~\ref{AAB}. This phenomenon indicates that the representations learned by deep learning models are discontinuous in semantic space. Ideally, derived samples with the same semantic information should be located in the neighborhood of the original samples, but they are often mapped far from the original samples in the model's output space. Specifically, semantic continuity means that when the semantic change in the input is minimal, the model's output should exhibit only minor changes. However, within the semantic neighborhood of a sample, there are still numerous instances that confuse the models or lead to incorrect classifications. This issue arises because deep learning models do not fully utilize the training data.

We believe that improving the efficiency of utilizing the semantic neighborhood of data is a feasible path to enhancing training. In deep learning algorithms, the utilization of the semantic neighborhood can be divided into two stages:

(1) Semantic Neighborhood Sampling:
Data augmentation~\cite{DBLP:journals/jiis/BaeckeP11, 
Buslaev2020AlbumentationsFA, Mikoajczyk2018DataAF, DBLP:journals/jbd/ShortenK19} and adversarial training~\cite{kurakin2016adversarial, Ian2017Explaining, nicholas2017towards} are the two most common neighborhood sampling methods. Data augmentation greatly alleviates overfitting and enhances the generalization of deep neural networks. Adversarial training, proposed to address adversarial attacks, involves injecting adversarial samples into the training set so that the model can simultaneously fit both the original and adversarial samples. While data augmentation improves model generalization to a certain extent, adversarial training enhances robustness. Recent studies have shown that adversarial training can bring many additional benefits~\cite{Tsipras2019RobustnessMB, Zhang2019InterpretingAT, Noack2019DoesIO, Santurkar2019ComputerVW, 
Engstrom2019LearningPR, Engstrom2019AdversarialRA, Goyal2020DROCCDR, Salehi2020ARAEAR, Salman2020DoAR, 
Utrera2020AdversariallyTrainedDN, Terzi2020AdversarialTR}, demonstrating that reasonable use of neighborhood samples can significantly improve model performance.

(2) Semantic Continuity: 
Neighborhood sampling generates many semantically similar images, but the model outputs for these samples are not always consistent. We define semantic neighborhood continuity as the property where slight semantic perturbations lead to only small changes in model output. Some initial work has utilized sample neighborhood continuity to enhance performance. For instance, leveraging neighborhood continuity has improved the effectiveness of image similarity searches~\cite{Zheng2016ImprovingTR}. Adversarial logits pairing, which also utilizes neighborhood continuity, significantly enhances the effectiveness of adversarial training~\cite{Kannan2018AdversarialLP}.
Another example is contrastive learning algorithms, which also leverage Neighborhood Continuity to achieve better pre-training performance on limited data.

The goal of this work is to explore in depth how semantic continuity affects models. We believe that the research on semantic neighborhood continuity is still in its very early stages. Semantic neighborhood sampling is merely a preliminary method to improve model generalization by using neighborhood data. Models trained with semantic neighborhood sampling do not learn the relationship between the original data and the data derived from this process; they simply treat the newly generated data as unrelated new samples. Existing training methods do not thoroughly explain why neighborhood continuity affects models, nor do they uncover the potential value of neighborhood continuity across various fields. Therefore, enhancing the semantic continuity of deep learning models is an insightful and promising endeavor.
This work addresses the issue of semantic continuity in deep learning models, presenting the following key contributions:
\begin{itemize}
\item We contextualize the problem of semantic discontinuity within existing deep learning models and propose a straightforward semantic continuity constraint. This constraint aims to facilitate smooth gradients and foster semantic-oriented features.
\item We conducted a series of experiments to evaluate the effectiveness of the proposed constraint. Our results demonstrate its efficacy in enhancing semantic continuity and show that models with improved semantic continuity exhibit notable improvements in robustness, interpretability, model transferability, and data bias issues.
\end{itemize}

\begin{figure*}[t]
   \begin{center}
      \includegraphics[width=0.28\linewidth]{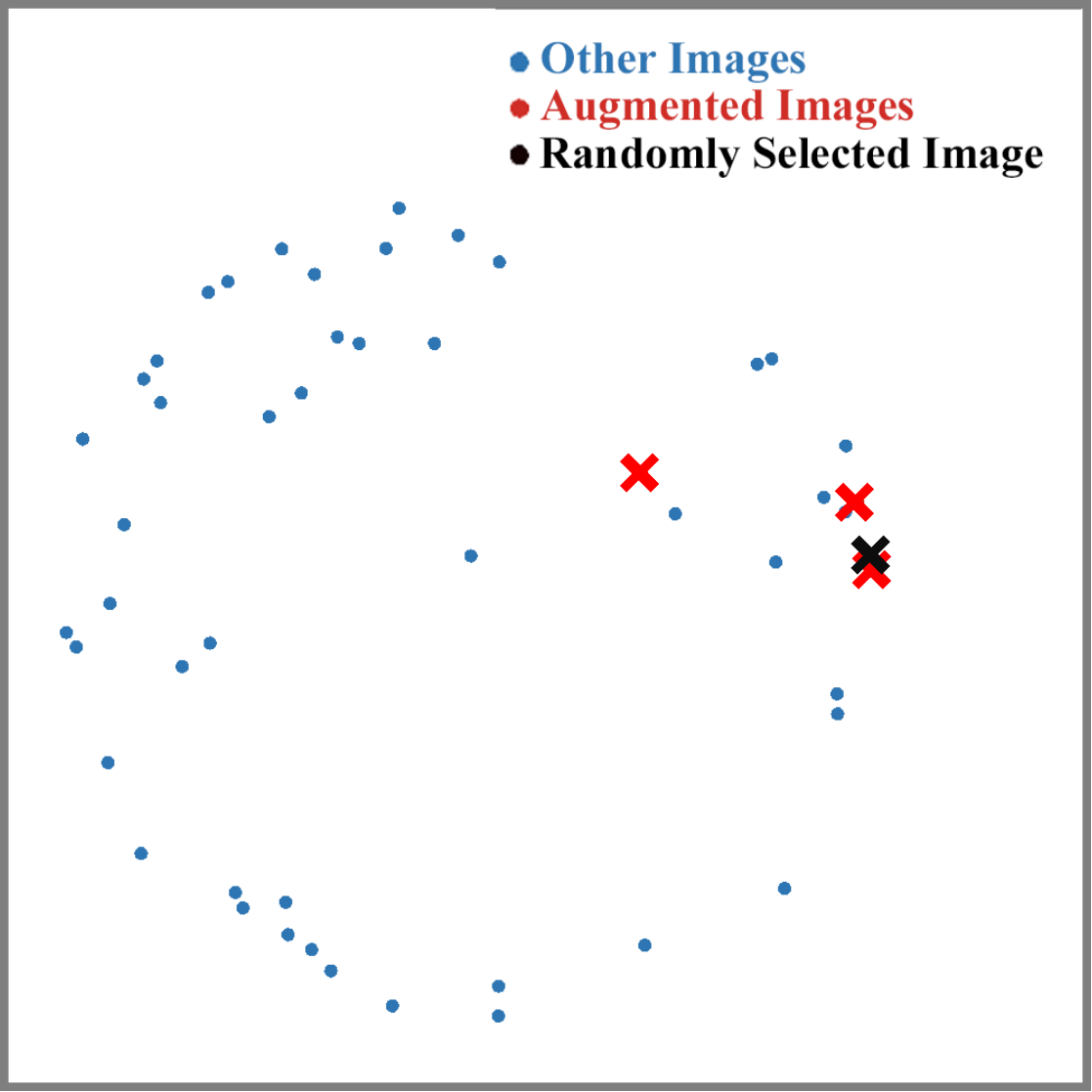} \hspace{10mm}
      \includegraphics[width=0.28\linewidth]{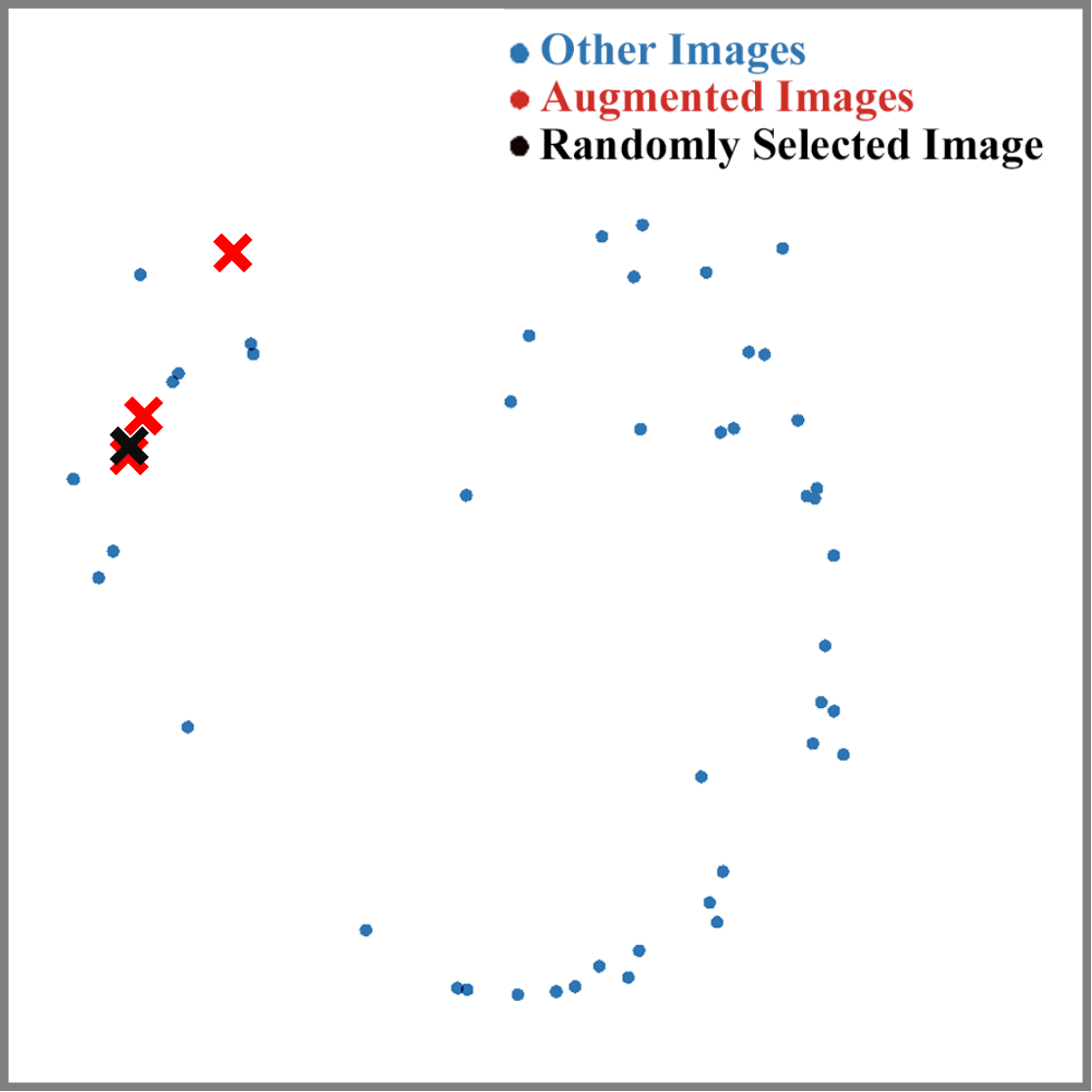} \hspace{10mm}
      \includegraphics[width=0.28\linewidth]{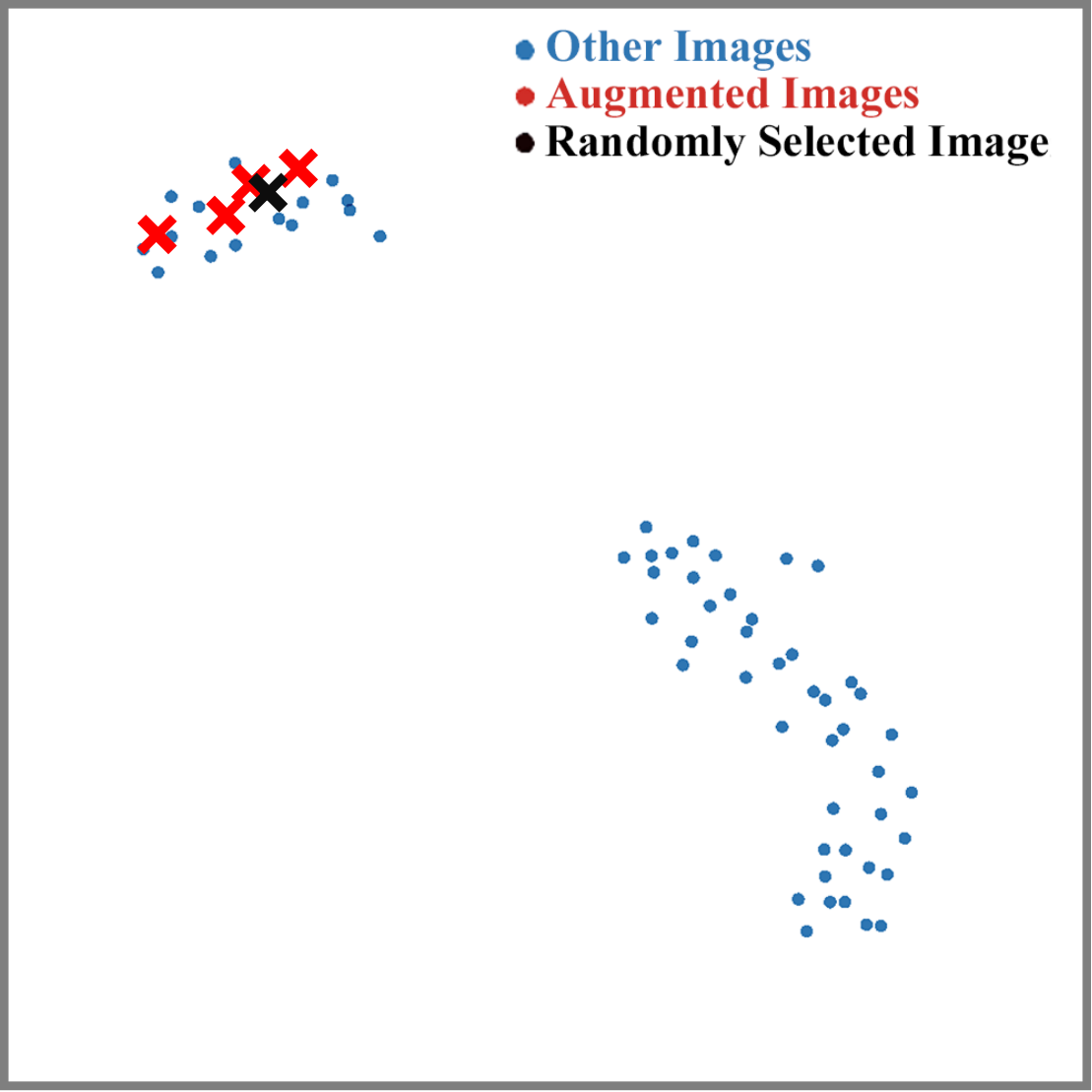}
   \end{center}
      \caption{
         Two-dimensional visualization of the representation of all samples in a specific category of the ImageNet Validation set. Black dots represent randomly selected images, while red dots indicate the representation positions of data-augmented images derived from one of the selected images. The left figure illustrates the result obtained using PCA, the middle one displays the TSNE result, and the right one showcases the outcome of UMAP.
         }
   \label{TSNE}
\end{figure*}

\section{Related Works}

\subsection{Training Paradigms}
The objectives of deep learning training have undergone several iterations, evolving from the initial supervised and unsupervised learning to other paradigms like reinforcement learning~\cite{wang2022deep} and meta-learning~\cite{hospedales2021meta}, each tailored to different scenarios and tasks. These paradigms aim to maximize the efficiency of knowledge acquisition given the limited representational capacity of models. Supervised learning remains the most effective training paradigm. However, the challenge of annotating vast amounts of data has led to the development of self-supervised and contrastive learning algorithms~\cite{jaiswal2020survey,liu2021self} in recent years. These methods enable efficient training on unlabeled data. The success of large language models~\cite{chang2024survey} and vision pretraining models~\cite{chen2023vlp} attests to the efficacy of these paradigms. A similar training concept can be traced back to face recognition algorithms utilizing triplet loss~\cite{schroff2015facenet}, where learning an effective representation space involves comparing pairs of similar and dissimilar entities. Adversarial training~\cite{qian2022survey} has garnered attention due to the discovery of adversarial examples. This paradigm significantly enhances model robustness. Adversarial Logit Pairing (ALP)~\cite{Kannan2018AdversarialLP} is a novel adversarial training method that leverages contrastive ideas to further improve robustness. Throughout the evolution of these training paradigms, the efficiency of data utilization has been gradually enhanced, leading to models that learn representation spaces with more human-like semantic characteristics. Therefore, exploring research directions that leverage semantic information is both meaningful and impactful.

\subsection{Trustworthy Machine Learning}

As deep learning models surpass human performance in various tasks, it becomes clear that accuracy alone is insufficient. Models must also exhibit generalization~\cite{HALLAJI2023110384,YI2022108831}, robustness~\cite{WANG2021107141}, interpretability, transferability~\cite{KE2024111909,DING2021107102}, and fairness~\cite{ZHANG2023110777} to ensure trustworthiness in their outputs~\cite{eshete2021making,DBLP:journals/isci/WangZLWZ23}. This is crucial to avoid scenarios like the infamous "tanks and cloudy skies" story~\cite{verma2020counterfactual}, where models exploit spurious correlations in the data. Thus, model evaluation should be multifaceted, not solely focused on task performance accuracy but also on whether the model achieves its objectives without relying on data-specific artifacts that inflate its perceived accuracy. Researchers have addressed issues related to model robustness, such as adversarial defense, backdoor attack defense, and model theft~\cite{li2022defending}. In terms of interpretability, efforts have been made to use model behavior and gradient information to analyze what the model attends to in the input. Additionally, substantial work has been done to correct and discuss biases learned from data, ensuring that model outputs align with fundamental societal ethics and moral standards. This holistic approach to model evaluation and improvement is essential for the development of reliable and trustworthy machine learning systems.

\section{Method}\label{sec31}

\subsection{Definition}

Before our experiments~\footnote{In this part, we use Resnet50 to conduct experiments in the val set of ImageNet2012.}, we first define some key items to be used in this paper:
\begin{enumerate}
   \item Image Representation: In this work, the activation value of the Logits layer of the 
   model is specified as the sample representation. 
   $F(\cdot)$ is the function to obtain the image representation.
   \item Representation Distance: We define Logits Difference ($LD$) as the distance vector between image representations.
   \begin{equation}
      LD(x, x') = F(x) - F(x')
   \end{equation}
   \item We define Distance Score($DS$) as the distance between the two representations.
   \begin{equation}
      DS(x, x') = ||LD(x, x')||_2^2
   \end{equation}
\end{enumerate}

\subsection{Problem Description}\label{SEC1}

Previous research~\cite{nicholas2017towards} has shown that adversarial samples exacerbate significant semantic discontinuity issues. This section presents data analysis results to substantiate that semantic discontinuity is prevalent, extending beyond adversarial scenarios to real-world normal samples subjected to various types of non-semantic perturbations.
Some traditional data augmentation methods can be considered non-semantic perturbations, including adjustments to brightness, contrast, saturation, and hue. In our experiment, we selected an image \( x_A \) from the ImageNet validation set and randomly applied one of the four non-semantic perturbations to generate the data-augmented sample \( x_{A}' \). As shown in Fig~\ref{AAB}, it is easy to find samples \( x_B \) in the same category as \( x_A \), which satisfy \( DS(x_A, x_{A}') > DS(x_A, x_B) \). This demonstrates that the issue of semantic discontinuity is not confined to adversarial samples. During the ordinary training process, the model did not learn the relationship between the sample pairs, but merely fit the two types of samples independently.

We further analyze this phenomenon by visualizing the derived representations. We randomly selected an image \( x_{A} \) from the ImageNet validation set and applied the four data augmentation operations, each modifying one of the attributes of \( x_{A} \). We then combined the original image \( x_{A} \), the four augmented images \(\{x_{A_{1}}', x_{A_{2}}', x_{A_{3}}', x_{A_{4}}'\}\), and the remaining images of the same category to form a new set \(\mathcal{I}_{visualize}\). This set was fed into ResNet50 to obtain the image representations. Finally, we visualized the image representations of \(\mathcal{I}_{visualize}\) using PCA, TSNE~\cite{Maaten2008VisualizingDU}, and UMAP~\cite{DBLP:journals/jossw/McInnesHSG18}. 
As shown in Fig.~\ref{TSNE}, all three visualization methods reveal that the perturbed samples are not the closest to the original sample in the representation space. This indicates that during the representation learning process, deep models generally fail to maintain the correspondence between the original image and its perturbed versions. This further demonstrates the presence of semantic discontinuity in normal image samples.

\subsection{Semantic Continuity Constraints}

\begin{figure*}[t]
   \begin{center}
      \includegraphics[width=0.9\linewidth]{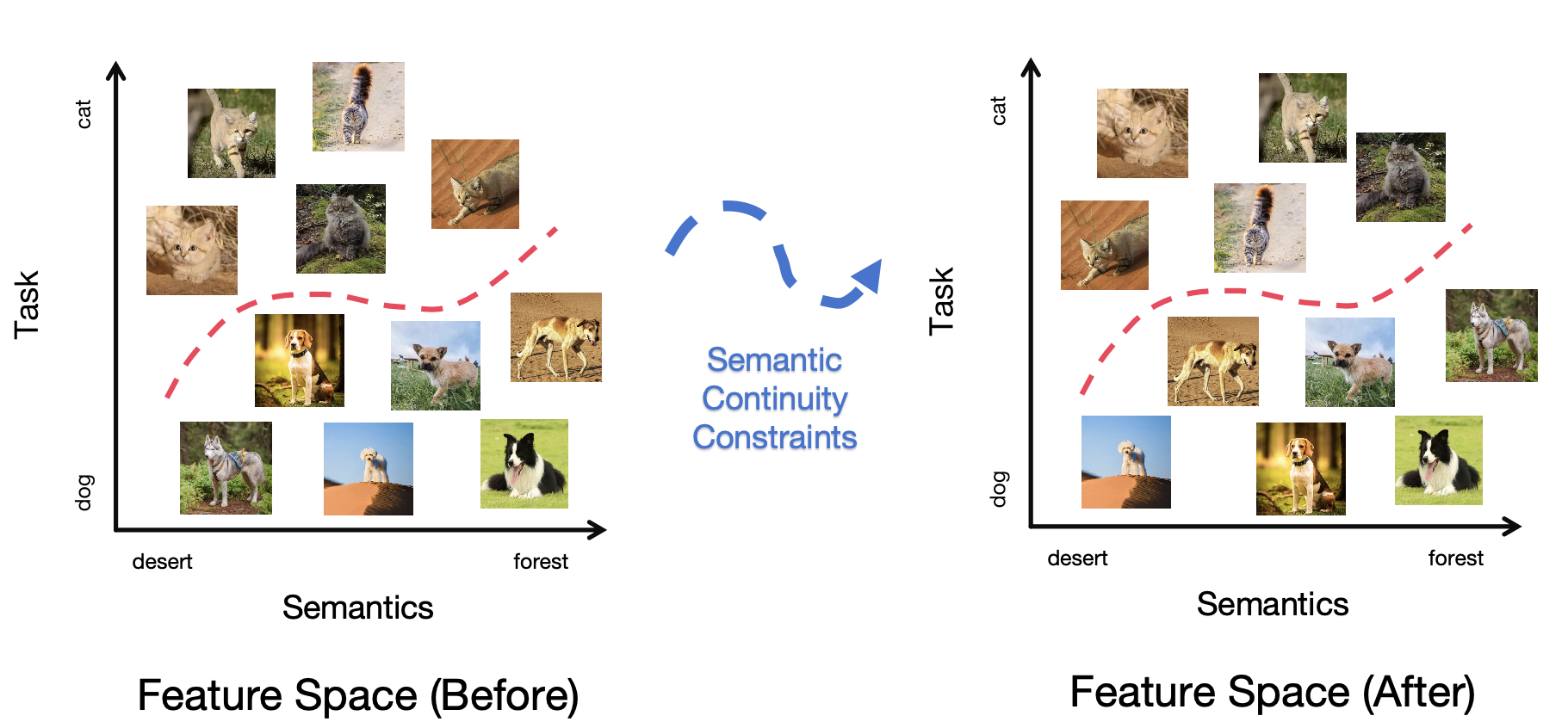}
   \end{center}
      \caption{
         We aim to build constraints based on non-semantic transforms, guiding the model to focus on semantic information while completing tasks, thereby improving the model's trustworthiness metrics. In the figure, the left side shows the feature space of standard training, and the right side shows the feature space after applying the semantic continuity constraint. For a cat-dog classification task, we use a desert-forest analogy to illustrate the desired effect we want the model to achieve. In the original training mode, the model only focused on the classification accuracy. We hope that under semantic constraints, the model can also pay attention to semantic information at the same time. For example, the semantic information related to the desert forest in the example is also correctly transferred.
         }
   \label{framework}
\end{figure*}

The phenomenon of semantic discontinuity in deep learning models contradicts human intuition. Ideally, deep models should interpret data in a similar way as humans. To achieve this, we introduce the semantic continuity constraint. We analyze its benefits for model learning and performance. The desired effect of the semantic continuity constraint can be intuitively understood from Fig.~\ref{framework}.

We previously observed that the representation distance ($DS$) between image pairs becomes unreasonably large under changes that do not affect the image's semantic information, such as data augmentation and adversarial attacks. Therefore, we intuitively use $DS$ to constrain the model and improve semantic continuity.
To calculate $DS$, semantically identical data pairs \( x \) and \( x' \) are required, where \( x \) refers to the original image and \( x' \) refers to the sample generated by non-semantic perturbation. Here, we detail the process of generating \( x' \):

\begin{equation}
   x' = P(x),
\end{equation}
where $P(\cdot)$ refers to the non-semantic perturbation: data augmentation or adversarial attack.

Based on the above discussion, we design a semantic continuity constraint method based on the Logits layer and choose data augmentation and adversarial attacks as tools to generate non-semantic perturbations.
\begin{equation}
   Loss_{continuity} = DS(x, x')
\end{equation}

Total loss can be defined as:
\begin{equation}
   Loss = Loss + \alpha * Loss_{continuity}
\end{equation}
where $\alpha$ is the weight of $Loss_{continuity}$.

It is noted that previous neighborhood continuity methods~\cite{Zheng2016ImprovingTR, Kannan2018AdversarialLP} can be seen as special cases of the semantic constraint. 
Here we give an explanation based on higher-order Taylor decomposition, 
elaborate on why the semantic continuity constraint can bring gains to the model.

Through the second-order Taylor decomposition, the previously defined $LD$ can be transformed into:
\begin{equation}
   LD(x,x')=F(x)-F(x')=\frac{\partial F}{\partial x}\Delta x + \Delta x^T \frac{\partial^2 F}{\partial x^2} \Delta x + R
\end{equation}
where $R$ is the remainder of the Taylor formula.

At present, data augmentation and adversarial attacks can meet the premise of not changing the semantic information.
In adversarial training, $\Delta x$ can be approximately expressed as: 
\begin{equation}
   \Delta x= \frac{\partial Loss}{\partial x} \alpha = \alpha \frac{\partial Loss}{\partial F}  \frac{\partial F}{\partial x}
   \label{equ_adv}
\end{equation}

Under the semantic continuity constraint, no matter what kind of non-semantic perturbation method is adopted, it will constrain $\frac{\partial F}{\partial x}$ and the higher-order derivative of the term to $x$. The physical meaning of those items represents the impact of a single pixel on the activation layer. 
The constraints on those items can limit the models' rely on individual pixels, and in turn encourage the models to learn more complex semantic features.
From Equ.~\ref{equ_adv} we can see that when adversarial training is used, the constraints on 
$\frac{\partial F}{\partial x}$ are further enhanced, so the adversarial training will further help the improvement of interpretable phenomena.

From the above analysis, it is evident that semantic continuity results in smoother input gradients. Smoothing the input gradient reduces the influence of irrelevant areas and individual pixels. Therefore, in theory, semantic continuity can enhance both the interpretability and robustness of the model. Improved robustness, in turn, benefits various applications such as transfer learning, image generation, and feature visualization~\cite{Zhang2019InterpretingAT, Noack2019DoesIO, Santurkar2019ComputerVW, 
Engstrom2019LearningPR, Engstrom2019AdversarialRA, Goyal2020DROCCDR, Salman2020DoAR, 
Utrera2020AdversariallyTrainedDN, Terzi2020AdversarialTR}.

\section{Experiment}

\subsection{Experiment Setting}

We conducted experiments on ImageNet2012 and Cifar100 datasets~\cite{Deng2009ImageNetAL, Russakovsky2015ImageNetLS, Krizhevsky2009LearningML}. Four ResNet50~\cite{He2016DeepRL} models were trained on ImageNet: a standard ResNet50, ResNet50C with semantic continuity constraints, ResNet50Adv for adversarial training, and ResNet50AdvC for combined semantic continuity and adversarial training. On Cifar100, ResNet18 was employed to train the four models mentioned above. We utilized the Adam optimizer~\cite{DBLP:journals/corr/KingmaB14} with a learning rate of 1e-5. PGD~\cite{nicholas2017towards} was employed to generate adversarial samples for adversarial training, with an attack step size of 2 pixels on Cifar100, $\epsilon=8$ pixels, and 10 iterations. On ImageNet, the attack step was 2 pixels, $\epsilon=16$ pixels, and it iterated for 10 rounds.

\subsection{Improvement On Semantic Continuity}

\begin{figure*}[!t]
   \centering
   \includegraphics[width=0.9\linewidth]{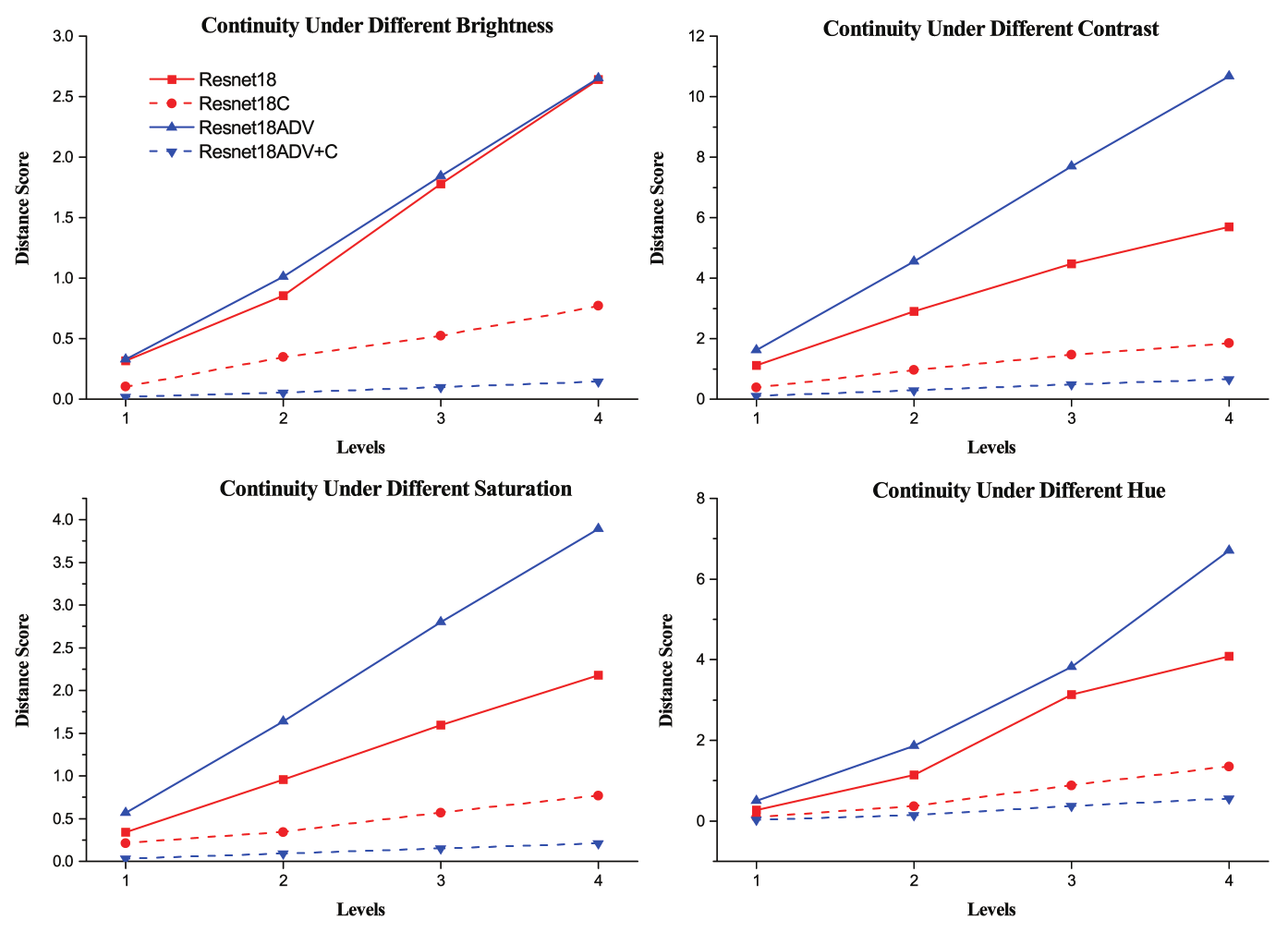}
   \caption{
      We measured the semantic continuity of the four models under non-semantic perturbations, examining the effects of Brightness, Contrast, Saturation, and Hue transformations. The results show that the semantic continuity of the models significantly improved after adding the semantic continuity constraint.
      }
   \label{fig:diff_result}
\end{figure*}

For classification tasks, the human decision-making process is insensitive to transformations in image brightness, contrast, saturation, and hue. In other words, these changes do not affect the semantic information of the image, but the output of a deep learning model can be influenced by these non-semantic perturbations. The semantic continuity loss constraint can effectively mitigate this deficiency in deep learning models. We tested the semantic continuity of four models—Resnet18, Resnet18C, Resnet18Adv, and Resnet18AdvC—under four augmentation methods: brightness, contrast, saturation, and hue. These augmentation methods were defined at four levels, as shown in Table~\ref{tab:param}. We measured the $DS$ of these models under these perturbations, as depicted in Fig.~\ref{fig:diff_result}. The results clearly show that the $DS$ of models with semantic continuity constraints is smaller than that of the original models. In other words, adding semantic continuity constraints reduces the influence of these four non-semantic perturbations, significantly improving the semantic continuity of the models.

\begin{table}[t]
   \centering
   \begin{tabular}{lllll}
       \hline
       Levels & Bright & Contrast & Saturation & Hue \\
       \hline
       1  & 16   & 1.25  & 1.25  & 0.1            \\
       2  & 32   & 1.50  & 1.50  & 0.2            \\
       3  & 48   & 1.75  & 1.75  & 0.3            \\
       4  & 64   & 2.0   & 2.0   & 0.4            \\         
       \hline       
   \end{tabular}
   \caption{For the non-semantic transformations used in the experiments, we employed four data augmentation algorithms to simulate them, each with four different intensities. The specific parameters corresponding to each intensity are shown in the table.}
   \label{tab:param}
\end{table}

\begin{figure}[t]
   \begin{center}
      \includegraphics[width=0.48\linewidth]{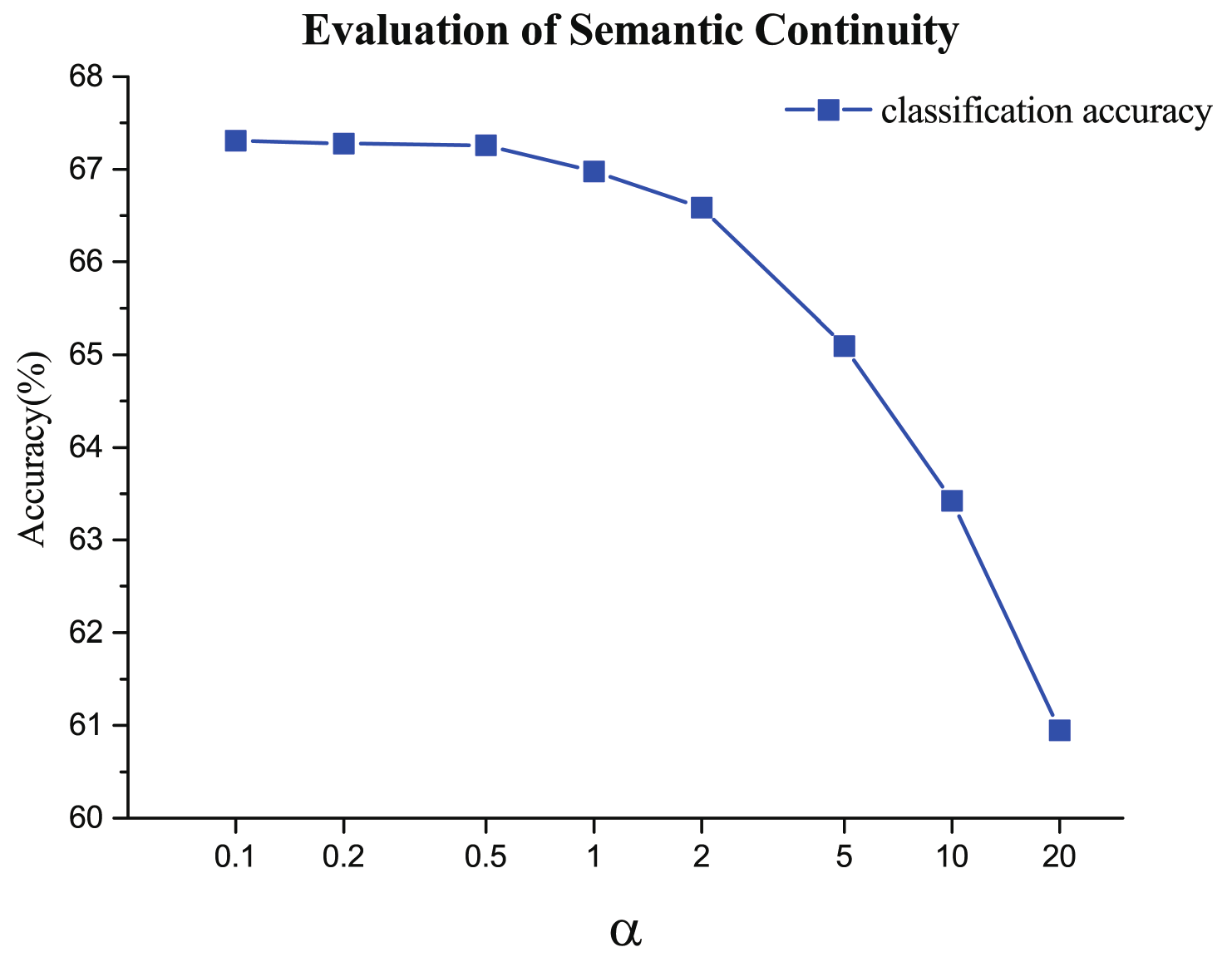}
      \includegraphics[width=0.48\linewidth]{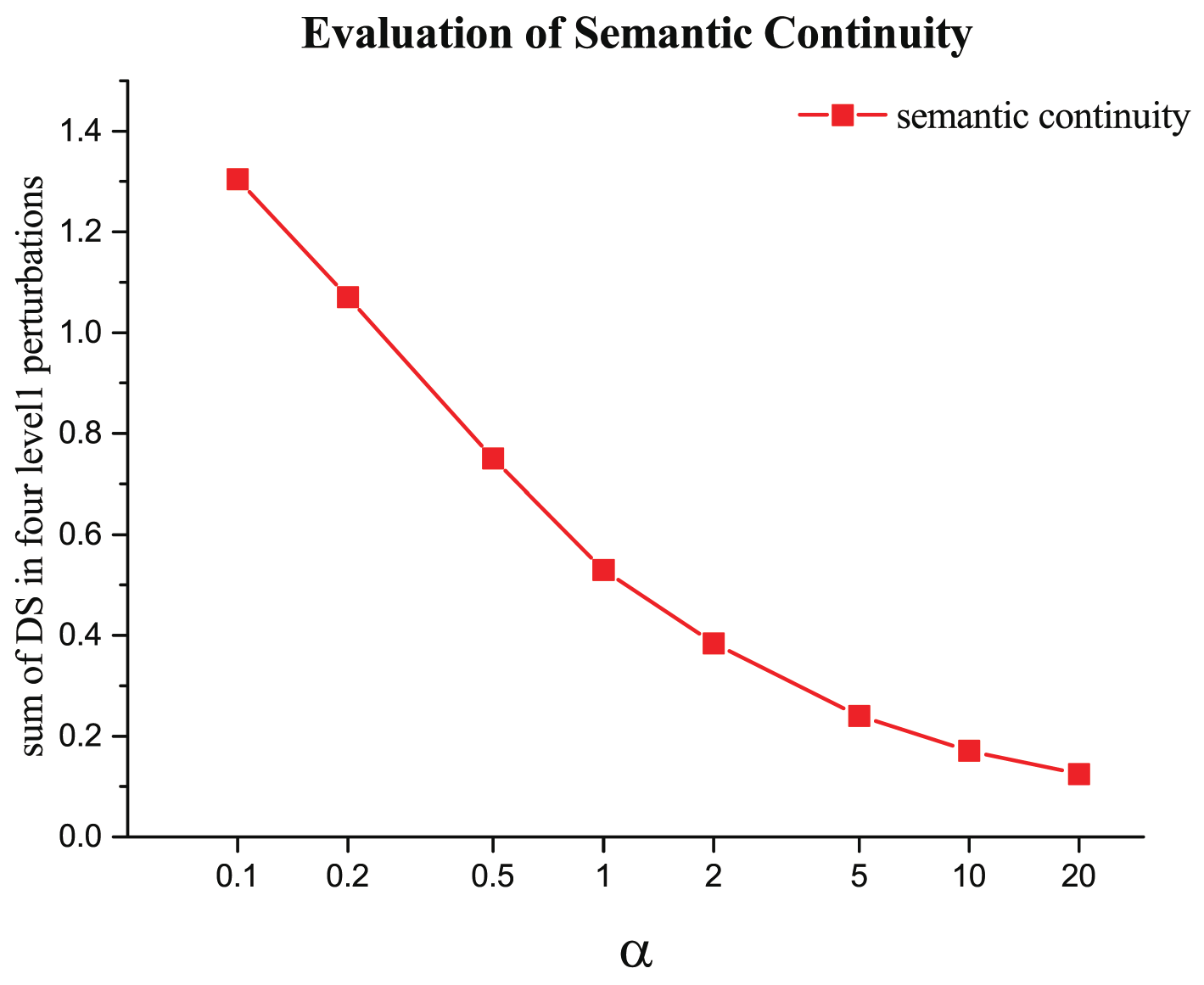}
   \end{center}
      \caption{The semantic continuity and classification accuracy under different settings of $\alpha$ are shown in the figures. The left figure illustrates the accuracy of models with varying $\alpha$ values, while the right figure displays the semantic continuity of models with different $\alpha$ values.}
   \label{pa}
\end{figure}

In the method part of this work, the semantic continuity's weight is designed as $\alpha$.
We measured the accuracy and semantic continuity of Resnet18 with different $\alpha$ in Cifar100. To measure the semantic continuity of the models, we calculated the sum of $DS$ values under four augmentation methods at level 1. As shown in Fig.~\ref{pa}, as $\alpha$ increases, the model's accuracy decreases slightly, but its semantic continuity improves significantly. Therefore, we ultimately set $\alpha=1.0$.

\subsection{Improving Semantic Continuity} \label{sec4}

In Section~\ref{sec31}, we demonstrated that the semantic continuity constraint could smooth the model's gradient. We also analyzed how a smoother gradient can reduce the model's reliance on individual pixels. To further investigate whether models constrained by semantic continuity utilize more semantic features, we designed three experiments.

\subsubsection{RGB Channels Translating}

\begin{figure}[t]
   \begin{center}
      \includegraphics[width=0.98\linewidth]{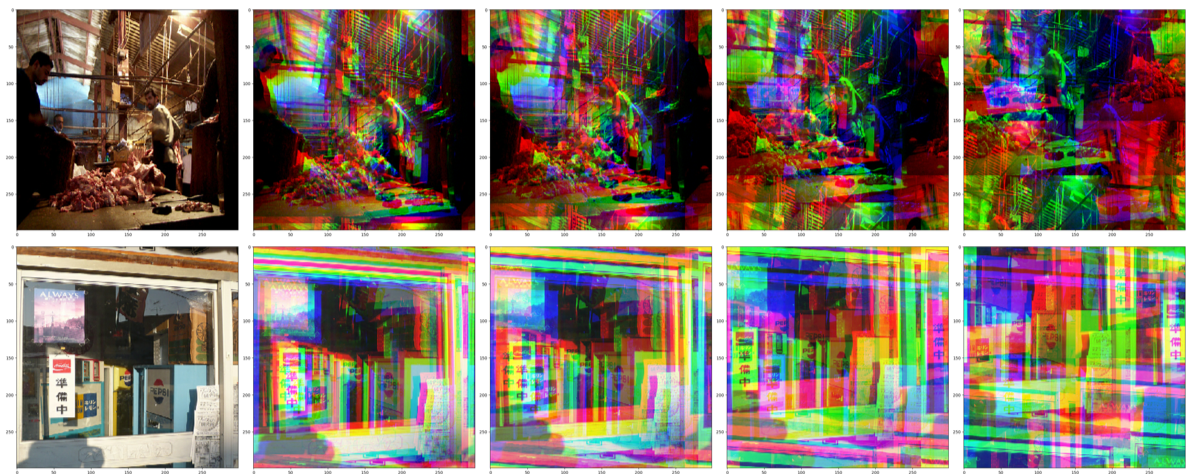}
   \end{center}
      \caption{
         The schematic image of RGB channel translation. From left to right: the original image, followed by images with translations of N = 1/32, 1/16, 1/8, and 1/4 of the side length.
         }
   \label{RGB_img}

   \begin{center}
      \includegraphics[width=0.9\linewidth]{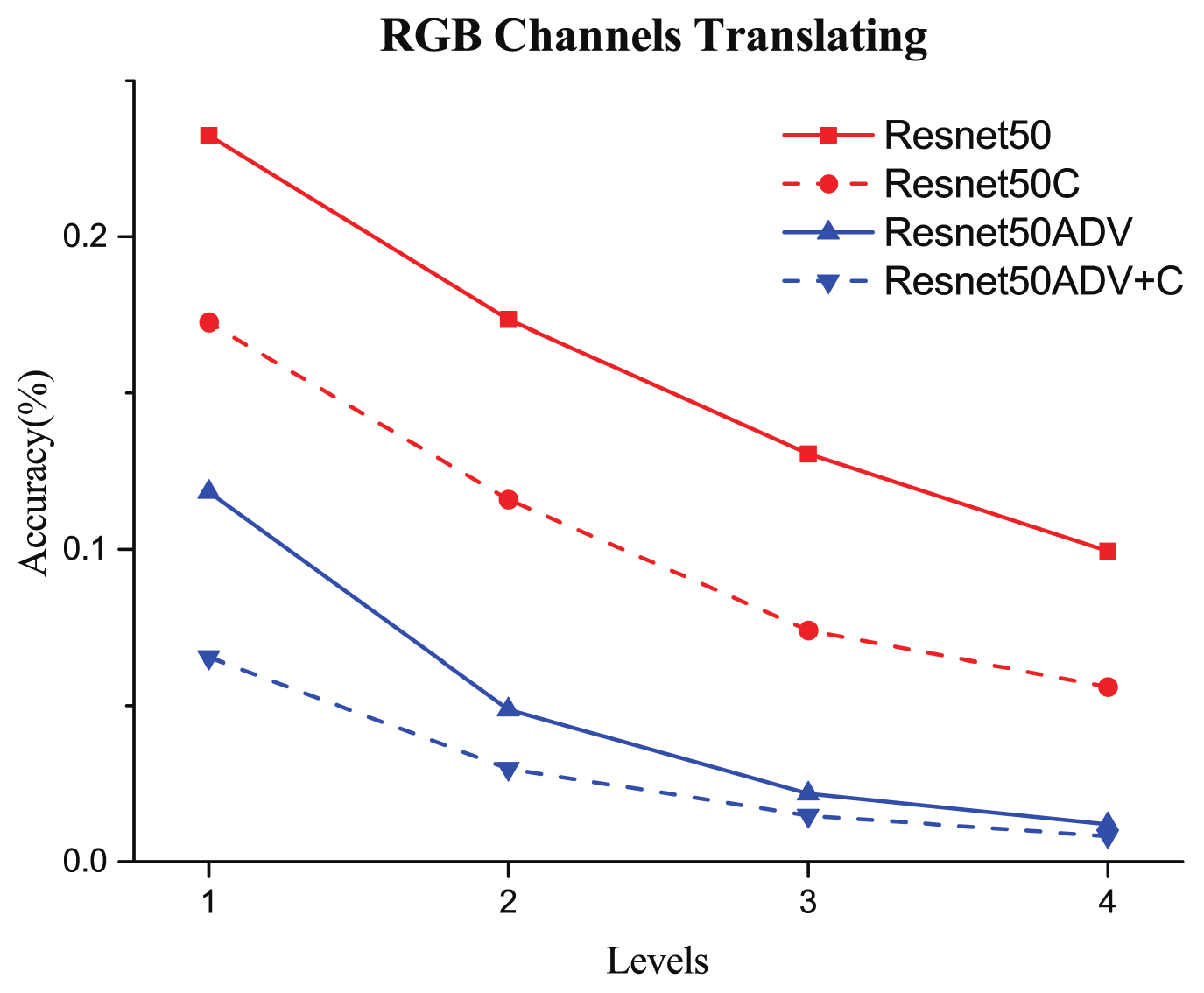}
   \end{center}
      \caption{
        We compared the accuracy of the four models on translated RGB channel images and found that the models with semantic continuity constraints are more sensitive to semantic perturbations. The X-axis is the perturbation level of the input image, and the y-axis is the accuracy in the test set.
         }
   \label{RBG_label}
\end{figure}

Given the translation invariance characteristic of convolutional neural networks, we devised a method to disrupt the semantic information of an image by translating its RGB channels, as illustrated in Fig.~\ref{RGB_img}. This approach maintains translation invariance within each channel but causes the image to lose its semantic integrity. Specifically, the $R$ channel is shifted to the lower right by $N$ pixels, and the $G$ channel by $N/2$ pixels. Simultaneously, the parts that move out of the image's edge are added to the upper left of each channel. As the translation distance increases, the image gradually loses its semantic content, becoming a color block that is difficult for humans to interpret. However, models are still capable of recognizing these types of images.

We conducted experiments to compare how models with and without semantic continuity constraints handle this transformed data. Table~\ref{RBG_label} presents the results, showing that models with stronger semantic continuity constraints rely less on unreasonable translational invariant information. Instead, these models focus more on identifying objects within the image, indicating a greater emphasis on semantic information.

\subsubsection{Randomization Center Area}

\begin{figure}[t]
   \begin{center}
      \includegraphics[width=0.98\linewidth]{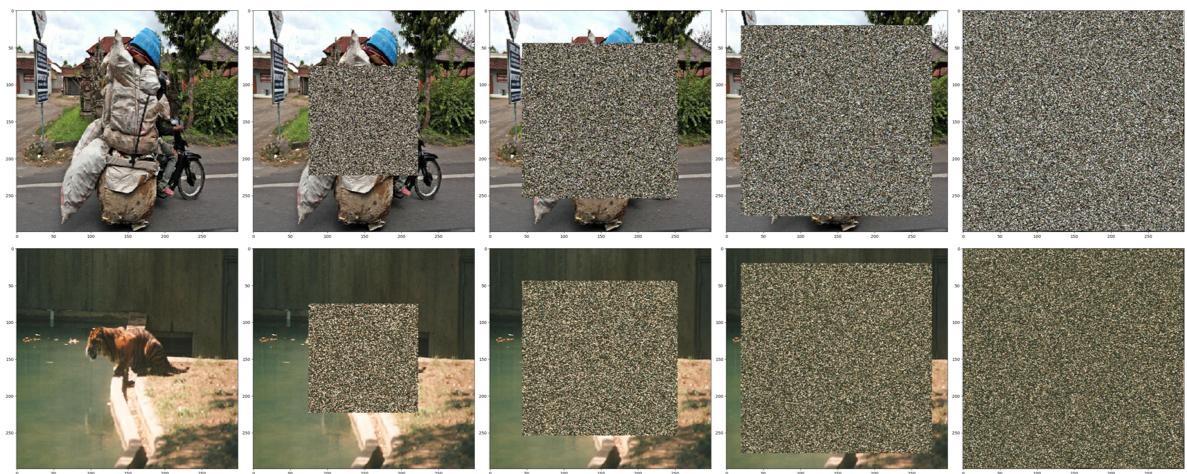}
   \end{center}
      \caption{
         On the left is the original image, followed by schematic images of the image after disrupting 25\%,50\%,75\%,100\% of the central part of the images.
         }
   \label{rand_img}
   \begin{center}
      \includegraphics[width=0.9\linewidth]{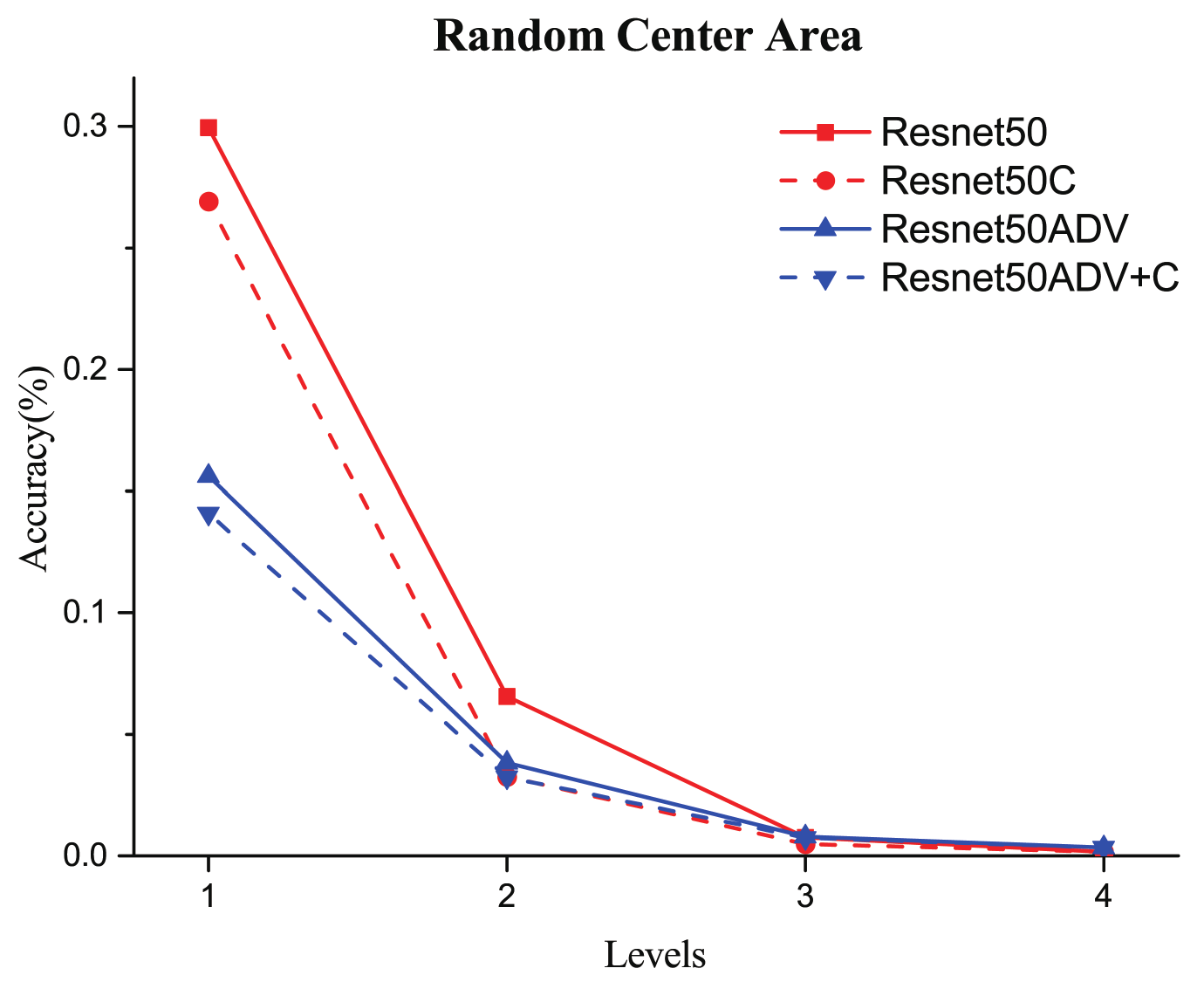}
   \end{center}
      \caption{
         We present the accuracy of the four models on images with a disrupted central area. Models with higher accuracy tend to use more task-unrelated information. It is evident that models with semantic continuity constraints rely more on task-related information.
         }
   \label{rand_label}
\end{figure}

We analyzed and counted images in the ImageNet validation set and found that over 95\% of images have task-related information centered. To test the models' reliance on this central information, we designed an experiment that disrupts the central pixels of the images, eliminating most task-related details, as illustrated in Fig.~\ref{rand_img}. The results in Table~\ref{rand_label} show that the accuracy of models with semantic continuity constraints is more affected by this disruption. This indicates that these models focus more on the foreground of the images and rely less on task-unrelated information.

\subsubsection{Sliding Puzzle}

\begin{figure}[!t]
   \begin{center}
      \includegraphics[width=0.98\linewidth]{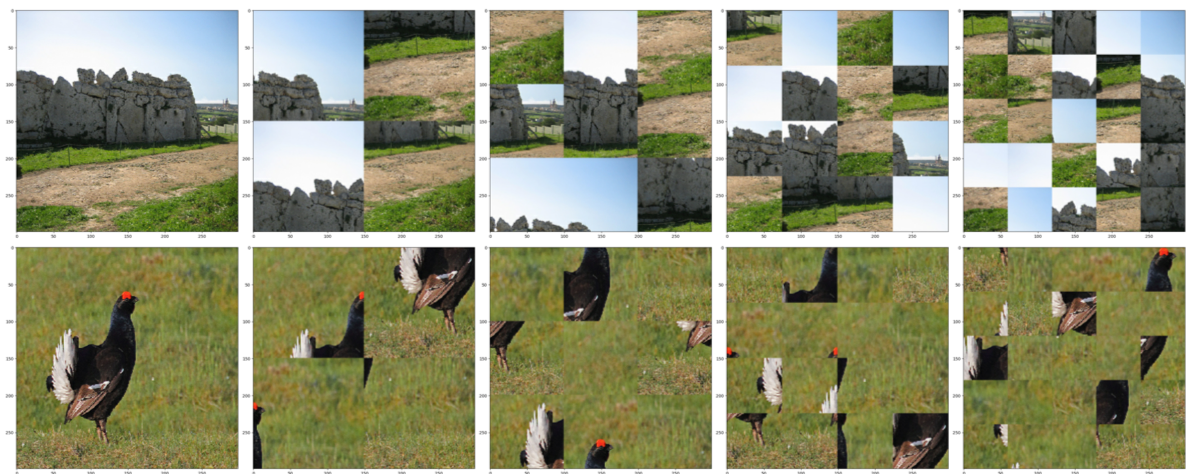}
   \end{center}
      \caption{
         Schematic illustration of images disrupted by random pixel blocks: the leftmost image is the original, followed by images disrupted into 4, 9, 16, and 25 blocks respectively.
         }
   \label{slide_img}
   \begin{center}
      \includegraphics[width=0.9\linewidth]{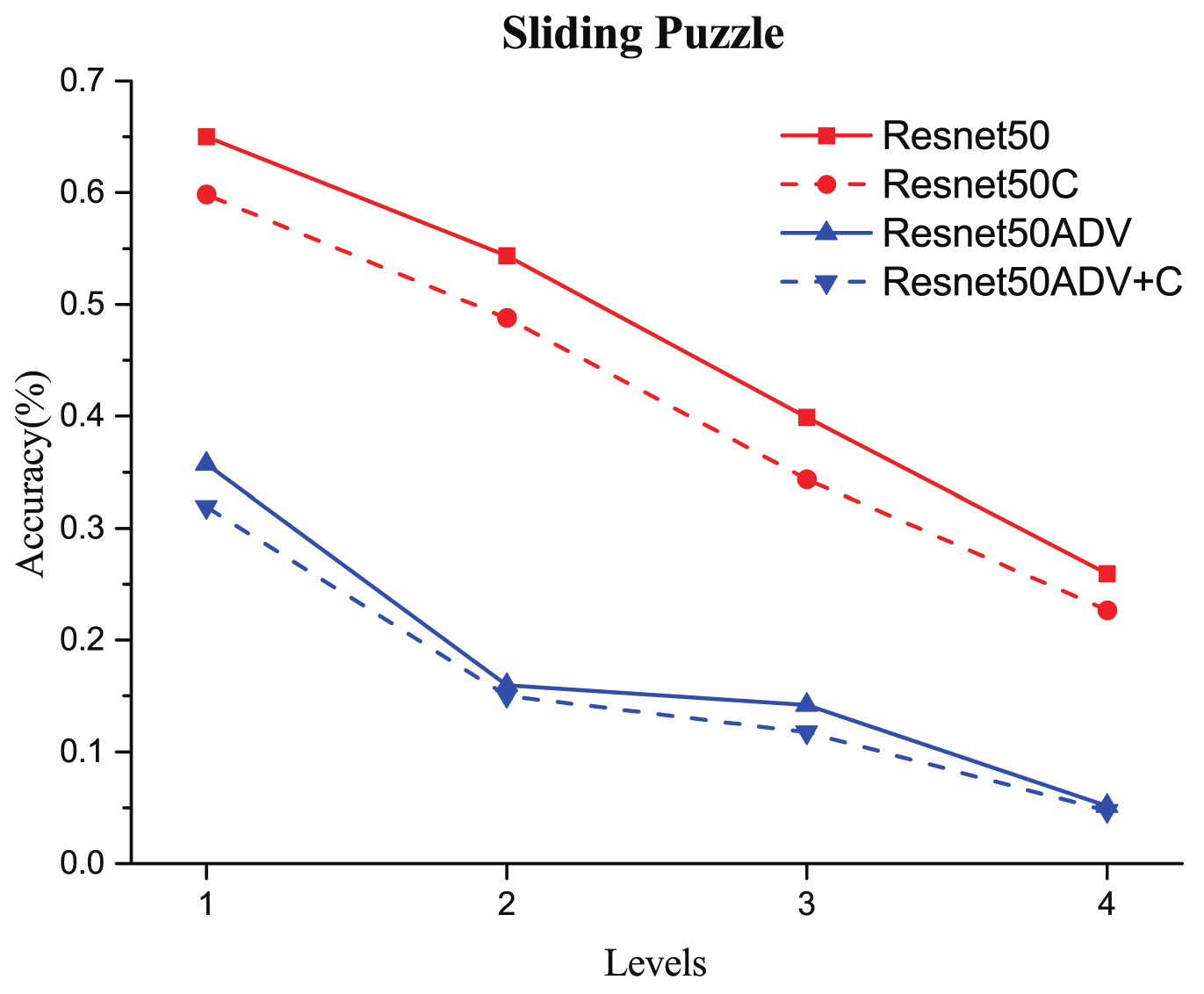}
   \end{center}
      \caption{
         As the number of disturbed image blocks increases, the image gradually loses semantic information. 
         The rapid decline in the classification accuracy of the re-spliced image of the model with continuity 
         constraints can explain its reduced use of non-semantic information.
      }
   \label{slide_label}
\end{figure}

Inspired by the work of Zhang et al.~\cite{Zhang2019InterpretingAT}. We designed an experiment, which divided each image into $N$ blocks and randomly shuffled their positions, as illustrated in Fig.~\ref{slide_img}. We then evaluated the classification accuracy of the models on these reassembled images. Despite the severe disruption causing the images to lose their original semantic information, the models still retained some classification capability, as shown in Table~\ref{slide_label}. This indicates that the models rely on non-semantic information, such as texture patterns.

The results of our comparative experiments demonstrate that models with semantic continuity constraints are less sensitive to non-semantic information and focus more on the semantic content of the images.

\subsection{Improving Adversarial Robustness} \label{sec_4adv}

The analysis in Sec.~\ref{sec31} shows that models with semantic continuity can reduce the impact of individual pixels, and since adversarial attacks primarily involve pixel-level perturbations, models with improved semantic continuity can also enhance adversarial robustness. To evaluate this, we used the PGD algorithm to generate white-box adversarial samples and assessed the model's robustness. For the ImageNet2012 and CIFAR-100 datasets, we set the step size to 2 pixels, iterating for 10 rounds with an attack setting of $\epsilon=8$ pixels.

We tested the classification accuracy on adversarial samples using four models derived from the ResNet50 and ResNet18 architectures. As shown in Table~\ref{adv_acc}, the robustness of adversarially trained models is significantly improved by adding a semantic continuity constraint. Additionally, models trained with the semantic continuity constraint, even without adversarial training, exhibit some defensive effects against adversarial samples. Therefore, incorporating semantic continuity substantially enhances model robustness.

\begin{table}[t]
   \begin{center}
      \begin{tabular}{|c|c|c|}
         \hline
         & clean accuracy & adv accuracy \\ \hline
         Resnet18 & 66.99\% & 9.34\% \\ \hline
         Resnet18C & \textbf{67.01\%} & 16.01\% \\ \hline
         Resnet18Adv & 63.80\% & 44.30\% \\ \hline
         Resnet18AdvC & 59.62\% & \textbf{54.04\%} \\ \hline
      \end{tabular}   
   \end{center}

   \begin{center}
      \begin{tabular}{|c|c|c|}
         \hline
         & clean accuracy & adv accuracy \\ \hline
         Resnet50 & \textbf{73.42\%} & 0.49\% \\ \hline
         Resnet50C & 72.01\% & 0.94\% \\ \hline
         Resnet50Adv & 55.43\% & 8.25\% \\ \hline
         Resnet50AdvC & 50.66\% & \textbf{32.55\%} \\ \hline
      \end{tabular}   
   \end{center}
   \caption{
      Comparison of the robustness of ordinary training and semantic-continuity-constrained training, \textbf{clean accuracy} 
      represents the classification accuracy on the original images, 
      and \textbf{adv accuracy} represents the classification accuracy on the adversarial samples.
      }
   \label{adv_acc}
\end{table}

\subsection{Improving Model Interpretability}

In Sec.~\ref{sec4}, we present three experiments demonstrating that the semantic continuity constraint increases models' reliance on semantic features and task-related information. Additionally, the semantic continuity constraint can mitigate the influence of individual pixels by smoothing the gradient. Combining these two points, we hypothesize that the semantic continuity constraint can also enhance interpretability.
To verify our hypothesis, we selected three interpretability algorithms—Integrated Gradients~\cite{DBLP:journals/corr/SundararajanTY16}, GradCAM~\cite{Selvaraju2019GradCAMVE}, and LIME~\cite{Ribeiro2016WhySI}—to observe how the semantic continuity constraint improves model interpretability.

\subsubsection{Integrated Gradients}

The Integrated Gradients Method~\cite{DBLP:journals/corr/SundararajanTY16} is an enhancement of the traditional Saliency Map~\cite{Kadir2004SaliencySA}, which is widely used in interpretability tasks today. It uses gradient integration to eliminate gradient saturation, allowing the gradients learned by the model to be more intuitively visualized.

\begin{figure}[t]
   \begin{minipage}[c]{0.48\linewidth}
      \centering
      \includegraphics[width=0.95\linewidth]{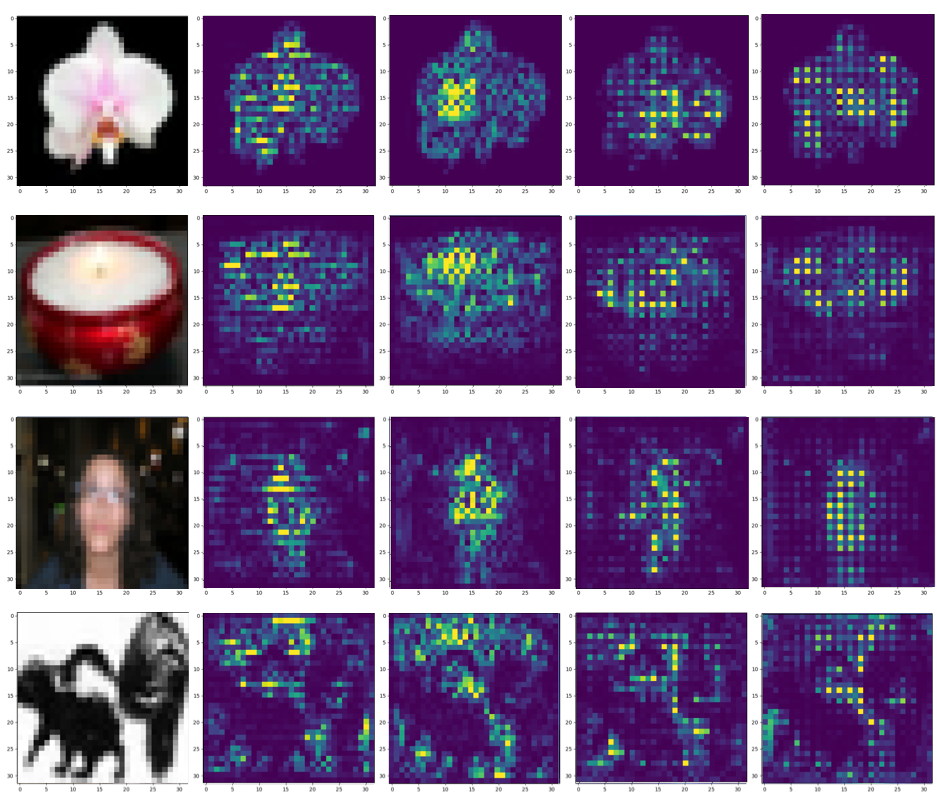}
        Cifar100(Resnet18)
   \end{minipage}
   \begin{minipage}[c]{0.48\linewidth}
      \centering
      \includegraphics[width=0.95\linewidth]{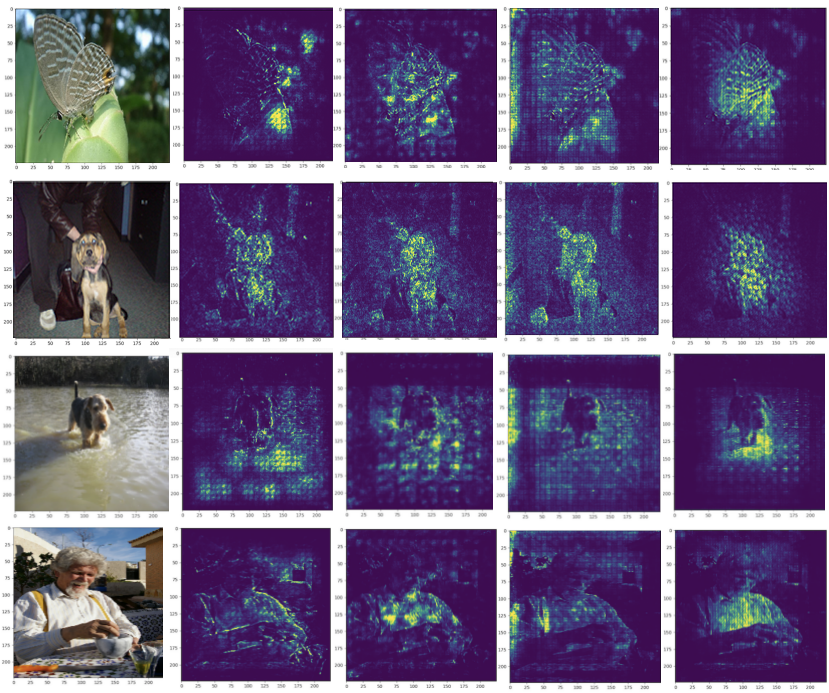}
        ImageNet(Resnet50)
   \end{minipage}
      \caption{
      The above figure is the Saliency Map of the different models we tested on the Cifar100 and Imagenet datasets.
      The first column is the original image and the second to fifth columns 
      are the gradient saliency maps of the common model, semantic continuity model, 
      robust model, and semantic continuity robust model.
      }
   \label{salency_img}
\end{figure}

Compared to the traditional Saliency Map, the Integrated Gradients Method offers improved visualization. We evaluated the performance of Resnet18 and Resnet50 network structures on the Cifar100 and ImageNet datasets, respectively. Fig.~\ref{salency_img} displays some representative interpretable images. Taking the third row of ImageNet results as an example, Resnet50C focuses more on the dog compared to Resnet50. Additionally, the gradient perturbations decrease significantly, resulting in smoother gradients. For Resnet50AdvC, this improvement is even more pronounced. From these results, it is evident that the addition of the semantic continuity constraint makes the gradients obtained by the model smoother and more focused on the recognized object, thereby greatly enhancing the model's interpretability.

\subsubsection{GradCAM}

\begin{figure}[t]
   \begin{minipage}[c]{0.48\linewidth}
    \centering
    \includegraphics[width=0.95\linewidth]{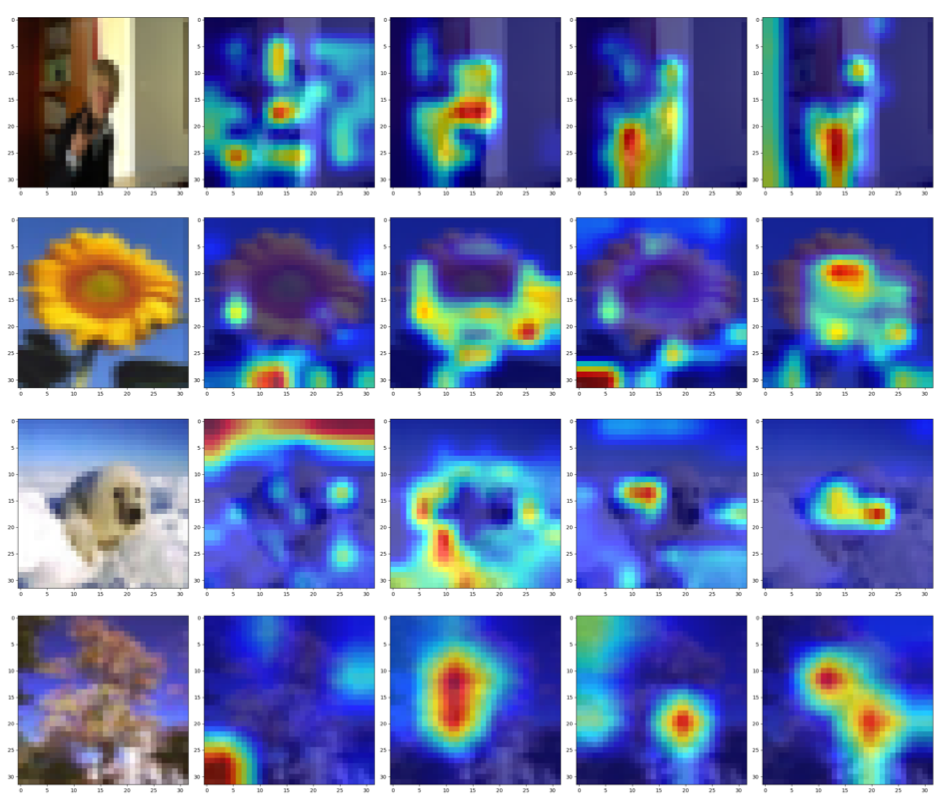}
        Cifar100(Resnet18)
   \end{minipage}
   \begin{minipage}[c]{0.48\linewidth}
    \centering
    \includegraphics[width=0.95\linewidth]{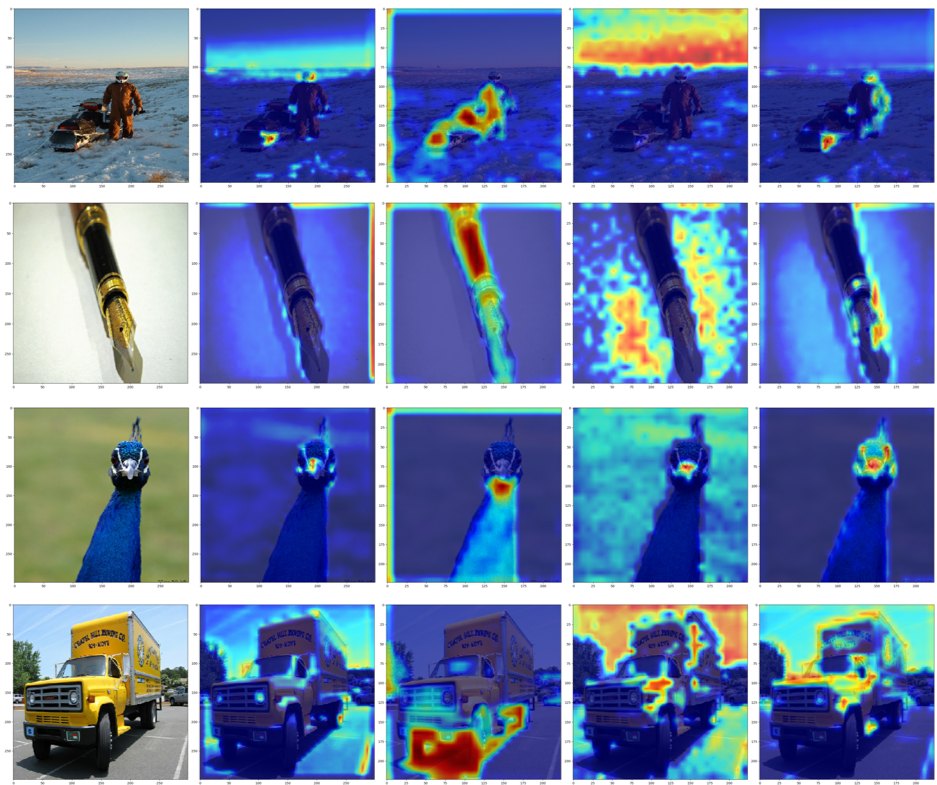}
        ImageNet(Resnet50)
   \end{minipage}
      \caption{
      The first column is the original image and the second to fifth columns 
      are the GradCAM results of the common model, semantic continuity model, 
      robust model, and semantic continuity robust model.
      }
   \label{gradcam_img}
\end{figure}

The GradCAM algorithm~\cite{Selvaraju2019GradCAMVE} utilizes gradient and activation information to visualize the model's regions of interest.
We employed GradCAM to observe the model's attention areas on images under different training constraints. The results display the activation maps of Resnet18 and Resnet50 on the Cifar100 and ImageNet datasets in Fig.~\ref{gradcam_img}. For instance, in the first row of ImageNet data, on the task of classifying snowmobiles, the Resnet50 model primarily focuses on the sky above the snow, whereas the Resnet50C model concentrates more on snowmobiles and people. This distinction becomes more pronounced with Resnet50Adv and Resnet50AdvC. The Resnet50Adv model's attention is more dispersed and distributed across the background, while Resnet50AdvC’s attention is more concentrated on the target object.

These findings demonstrate that the semantic continuity constraint refines the model's focus on task-related information, thereby enhancing its interpretability.

\subsubsection{LIME}

\begin{figure}[t]
   \begin{minipage}[c]{0.48\linewidth}
    \centering
    \includegraphics[width=0.95\linewidth]{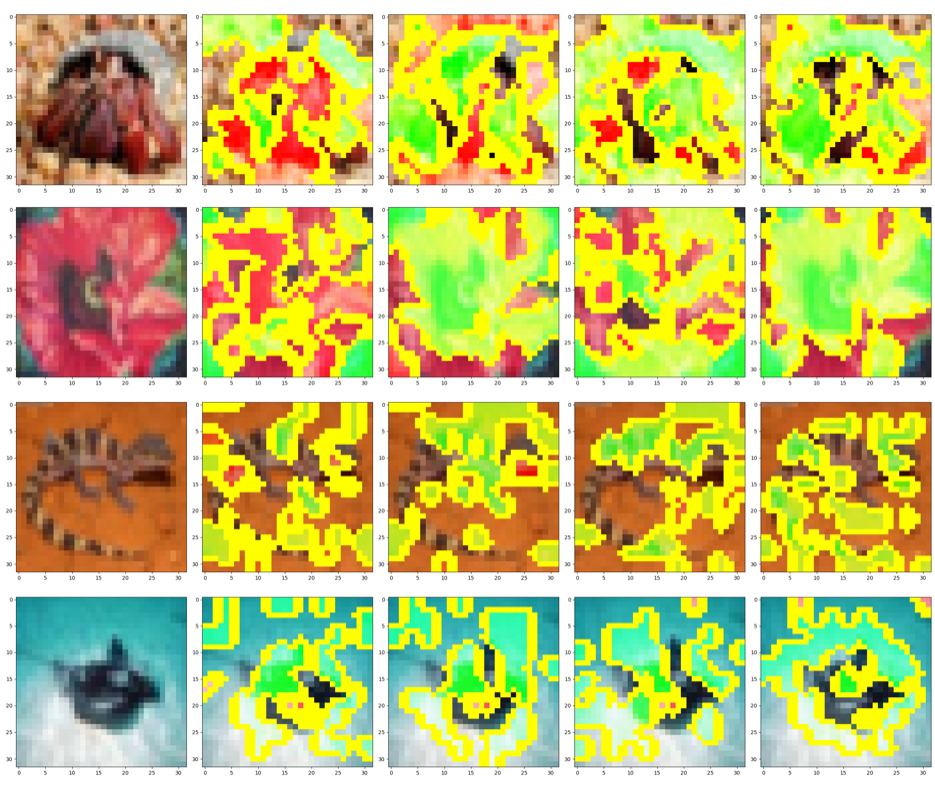}
        Cifar100(Resnet18)
   \end{minipage}
   \begin{minipage}[c]{0.48\linewidth}
      \centering
      \includegraphics[width=0.95\linewidth]{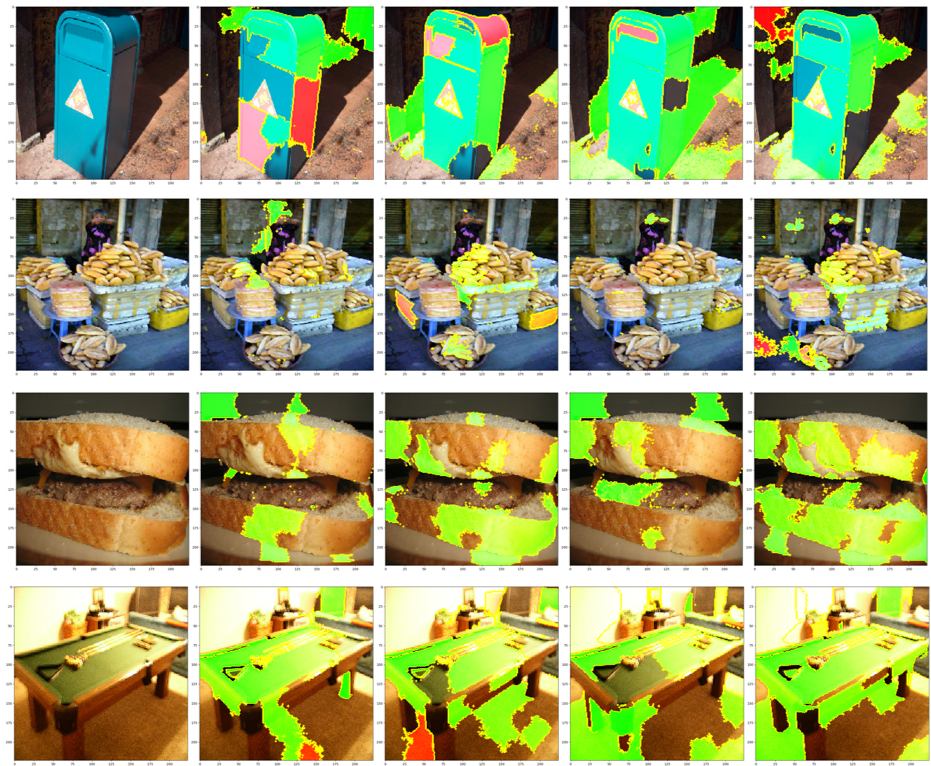}
        ImageNet(Resnet50)
   \end{minipage}
  \caption{
  The first column is the original image and the second to fifth columns 
  are the LIME results of the common model, semantic continuity model, 
  robust model, and semantic continuity robust model.
  }
    \label{lime_img}
\end{figure}

LIME is a method for analyzing image correlation based on output probability, providing an objective evaluation of model output. The results are shown in Fig.~\ref{lime_img}. Using the third row of ImageNet results as an example, we can analyze the performance of these models. Compared to Resnet50, Resnet50C is influenced by more areas within the bread, indicating a greater dependence on information within the bread. Similarly, it is evident that Resnet50AdvC relies more on the bread's information than Resnet50Adv does. This demonstrates that the addition of the semantic continuity constraint improves the LIME results. The model becomes more reliant on task-related information for its output, while the influence of task-unrelated information is diminished.

\subsection{Improving Transferability}

Robust models always transfer better~\cite{Salman2020DoAR, Utrera2020AdversariallyTrainedDN, Terzi2020AdversarialTR}. 
The experiments in Sec.~\ref{sec_4adv} prove that the semantic continuity constraint can improve the models' robustness. 
So we speculate the semantic-continuity-constrained models will have good transferability.
Therefore, we tested the transferability of these four models. 
We use the four Resnet18 models trained by Cifar100 to fine-tune the three data sets of Cifar10, SVHN, and MNIST. 
We find that the model with semantic continuity constraint has a higher 
accuracy rate, which shows that the model with semantic continuity constraint is helpful to model transferability.

\begin{table}[!t]
   \begin{center}
      \begin{tabular}{|c|c|c|c|}
         \hline
         Transfer & Cifar10 & SVHN & MNIST \\ \hline
         Resnet18 & 84.61\% & 94.25\% & 99.32\% \\ \hline
         Resnet18C & 84.78\% & 94.60\% & 99.37\% \\ \hline
         Resnet18Adv & 85.17\% & 94.66\% & 99.50\% \\ \hline
         Resnet18AdvC & \textbf{86.40\%} & \textbf{94.90\%} & \textbf{99.50\%} \\ \hline
      \end{tabular}   
   \end{center}
   \caption{
      The performance of normal training and semantic continuity constraint training in model transfer.
      }
   \label{transfer_label}
\end{table}

\subsection{Improving Fairness}

The data bias has attracted researchers more attention in recent years. 
The reason for the bias lies in the imbalance of training data. The model will 
learn the bias and imbalance in data and produce unreasonable 
outputs, such as gender discrimination. We believe that the key to 
solving this problem is to let the model minimize the use of task-unrelated 
information (such as gender) and focus more on task-related information. 
Previous experiments reveal that semantic continuity constraints can 
increase the model's use of semantic information and task-related information. 
Therefore, we speculate that semantic continuity constraint should also be helpful to the data 
bias.

We designed a Colorful MNIST dataset to verify our conjecture.
The background of 95\% data for the training set is red, and that of the left 5\% is blue; 
at the same time, the background of 95\% data for testing is blue, and that of the left 5\% is red.
We use normal training and semantic continuity training methods to fit the training set, and test their 
performance in the test set. The experimental results are shown in the Table~\ref{transfer_label}. 

It shows that the model with semantic continuity constraint uses less task-unrelated 
information (such as color), and the semantic continuity constraint is helpful in solving the problem of data bias.

\begin{figure}[!t]
   \begin{center}
      \includegraphics[width=0.98\linewidth]{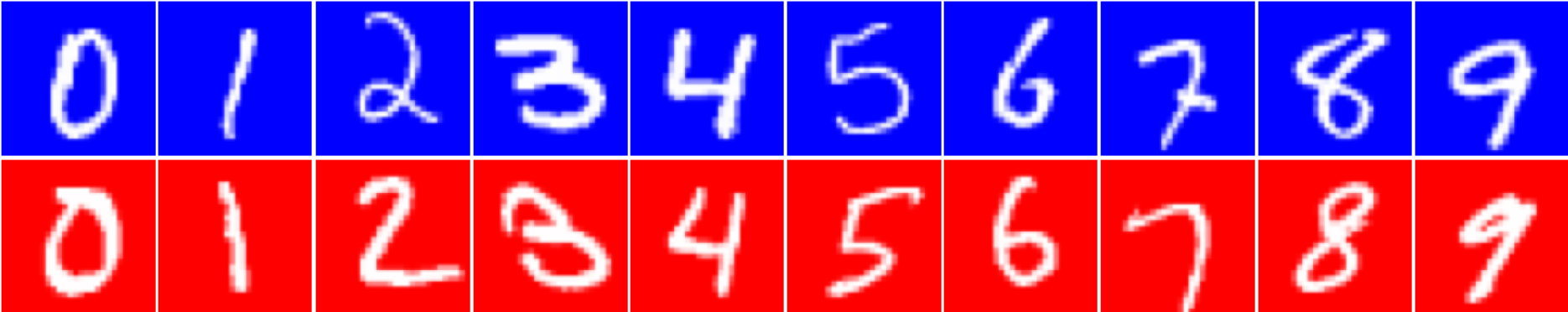}
   \end{center}
      \caption{
         Schematic image of samples in the Color MNIST dataset. We constructed a biased dataset based on MNIST to verify that the training model with semantic constraints has a better de-biasing effect.
         }
   \label{}
\end{figure}

\begin{table}[!h]
   \begin{center}
      \begin{tabular}{|c|c|}
         \hline
         & Color MNIST \\ \hline
         Resnet18 & 96.01\% \\ \hline
         Resnet18C & \textbf{97.54\%} \\ \hline
      \end{tabular}   
   \end{center}
   \caption{
      Test performance of ordinary training and semantic continuity constraint training on the Colorful MNIST dataset.
      }
\end{table}

\section{Conclusions}

Based on existing training paradigms and the direction of model training development, we abstractly propose a potential pathway to enhance data utilization efficiency, thereby improving model training effectiveness. Through experiments, we demonstrate that our proposed semantic continuity improves robustness, interpretability, model transferability, and fairness. We believe that semantic continuity brings model perception closer to human-like understanding. In the future, we plan to explore the application of this constraint in more complex scenarios and delve deeper into the implementation methods for non-semantic transformations. We hope our work can inspire researchers in the field of deep learning to design better training paradigms and model architectures.

\printcredits

\bibliographystyle{cas-model2-names}
\bibliography{egbib}

\begin{thebibliography}{49}
\expandafter\ifx\csname natexlab\endcsname\relax\def\natexlab#1{#1}\fi
\providecommand{\url}[1]{\texttt{#1}}
\providecommand{\href}[2]{#2}
\providecommand{\path}[1]{#1}
\providecommand{\DOIprefix}{doi:}
\providecommand{\ArXivprefix}{arXiv:}
\providecommand{\URLprefix}{URL: }
\providecommand{\Pubmedprefix}{pmid:}
\providecommand{\doi}[1]{\href{http://dx.doi.org/#1}{\path{#1}}}
\providecommand{\Pubmed}[1]{\href{pmid:#1}{\path{#1}}}
\providecommand{\bibinfo}[2]{#2}
\ifx\xfnm\relax \def\xfnm[#1]{\unskip,\space#1}\fi
%Type = Article
\bibitem[{Baecke and den Poel(2011)}]{DBLP:journals/jiis/BaeckeP11}
\bibinfo{author}{Baecke, P.}, \bibinfo{author}{den Poel, D.V.},
  \bibinfo{year}{2011}.
\newblock \bibinfo{title}{Data augmentation by predicting spending pleasure
  using commercially available external data}.
\newblock \bibinfo{journal}{J. Intell. Inf. Syst.} .
%Type = Inproceedings
\bibitem[{Benz et~al.(2020)Benz, Zhang, Karjauv and
  Kweon}]{Tsipras2019RobustnessMB}
\bibinfo{author}{Benz, P.}, \bibinfo{author}{Zhang, C.},
  \bibinfo{author}{Karjauv, A.}, \bibinfo{author}{Kweon, I.S.},
  \bibinfo{year}{2020}.
\newblock \bibinfo{title}{Robustness may be at odds with fairness: An empirical
  study on class-wise accuracy}, in: \bibinfo{booktitle}{NeurIPS}.
%Type = Article
\bibitem[{Buslaev et~al.(2020)Buslaev, Iglovikov, Khvedchenya, Parinov,
  Druzhinin and Kalinin}]{Buslaev2020AlbumentationsFA}
\bibinfo{author}{Buslaev, A.}, \bibinfo{author}{Iglovikov, V.I.},
  \bibinfo{author}{Khvedchenya, E.}, \bibinfo{author}{Parinov, A.},
  \bibinfo{author}{Druzhinin, M.}, \bibinfo{author}{Kalinin, A.A.},
  \bibinfo{year}{2020}.
\newblock \bibinfo{title}{Albumentations: Fast and flexible image
  augmentations}.
\newblock \bibinfo{journal}{Inf.} .
%Type = Inproceedings
\bibitem[{Carlini and Wagner(2017)}]{nicholas2017towards}
\bibinfo{author}{Carlini, N.}, \bibinfo{author}{Wagner, D.A.},
  \bibinfo{year}{2017}.
\newblock \bibinfo{title}{Towards evaluating the robustness of neural
  networks}, in: \bibinfo{booktitle}{{IEEE} Symposium on Security and Privacy}.
%Type = Article
\bibitem[{Chang et~al.(2024)Chang, Wang, Wang, Wu, Yang, Zhu, Chen, Yi, Wang,
  Wang et~al.}]{chang2024survey}
\bibinfo{author}{Chang, Y.}, \bibinfo{author}{Wang, X.}, \bibinfo{author}{Wang,
  J.}, \bibinfo{author}{Wu, Y.}, \bibinfo{author}{Yang, L.},
  \bibinfo{author}{Zhu, K.}, \bibinfo{author}{Chen, H.}, \bibinfo{author}{Yi,
  X.}, \bibinfo{author}{Wang, C.}, \bibinfo{author}{Wang, Y.}, et~al.,
  \bibinfo{year}{2024}.
\newblock \bibinfo{title}{A survey on evaluation of large language models}.
\newblock \bibinfo{journal}{ACM Transactions on Intelligent Systems and
  Technology} .
%Type = Article
\bibitem[{Chen et~al.(2023)Chen, Zhang, Han, Chen, Shi, Xu and
  Xu}]{chen2023vlp}
\bibinfo{author}{Chen, F.L.}, \bibinfo{author}{Zhang, D.Z.},
  \bibinfo{author}{Han, M.L.}, \bibinfo{author}{Chen, X.Y.},
  \bibinfo{author}{Shi, J.}, \bibinfo{author}{Xu, S.}, \bibinfo{author}{Xu,
  B.}, \bibinfo{year}{2023}.
\newblock \bibinfo{title}{Vlp: A survey on vision-language pre-training}.
\newblock \bibinfo{journal}{Machine Intelligence Research} .
%Type = Inproceedings
\bibitem[{Deng et~al.(2009)Deng, Dong, Socher, Li, Li and
  Fei{-}Fei}]{Deng2009ImageNetAL}
\bibinfo{author}{Deng, J.}, \bibinfo{author}{Dong, W.},
  \bibinfo{author}{Socher, R.}, \bibinfo{author}{Li, L.}, \bibinfo{author}{Li,
  K.}, \bibinfo{author}{Fei{-}Fei, L.}, \bibinfo{year}{2009}.
\newblock \bibinfo{title}{Imagenet: {A} large-scale hierarchical image
  database}, in: \bibinfo{booktitle}{CVPR}.
%Type = Article
\bibitem[{Ding et~al.(2021)Ding, Liu, Niu, Hu, Wang and Zhang}]{DING2021107102}
\bibinfo{author}{Ding, K.}, \bibinfo{author}{Liu, X.}, \bibinfo{author}{Niu,
  W.}, \bibinfo{author}{Hu, T.}, \bibinfo{author}{Wang, Y.},
  \bibinfo{author}{Zhang, X.}, \bibinfo{year}{2021}.
\newblock \bibinfo{title}{A low-query black-box adversarial attack based on
  transferability}.
\newblock \bibinfo{journal}{Knowledge-Based Systems} .
%Type = Article
\bibitem[{Engstrom et~al.(2019a)Engstrom, Ilyas, Santurkar, Tsipras, Tran and
  Madry}]{Engstrom2019AdversarialRA}
\bibinfo{author}{Engstrom, L.}, \bibinfo{author}{Ilyas, A.},
  \bibinfo{author}{Santurkar, S.}, \bibinfo{author}{Tsipras, D.},
  \bibinfo{author}{Tran, B.}, \bibinfo{author}{Madry, A.},
  \bibinfo{year}{2019}a.
\newblock \bibinfo{title}{Adversarial robustness as a prior for learned
  representations}.
\newblock \bibinfo{journal}{CoRR} .
%Type = Article
\bibitem[{Engstrom et~al.(2019b)Engstrom, Ilyas, Santurkar, Tsipras, Tran and
  Madry}]{Engstrom2019LearningPR}
\bibinfo{author}{Engstrom, L.}, \bibinfo{author}{Ilyas, A.},
  \bibinfo{author}{Santurkar, S.}, \bibinfo{author}{Tsipras, D.},
  \bibinfo{author}{Tran, B.}, \bibinfo{author}{Madry, A.},
  \bibinfo{year}{2019}b.
\newblock \bibinfo{title}{Learning perceptually-aligned representations via
  adversarial robustness}.
\newblock \bibinfo{journal}{CoRR} .
%Type = Article
\bibitem[{Eshete(2021)}]{eshete2021making}
\bibinfo{author}{Eshete, B.}, \bibinfo{year}{2021}.
\newblock \bibinfo{title}{Making machine learning trustworthy}.
\newblock \bibinfo{journal}{Science} .
%Type = Inproceedings
\bibitem[{Goodfellow et~al.(2015)Goodfellow, Shlens and
  Szegedy}]{Ian2017Explaining}
\bibinfo{author}{Goodfellow, I.J.}, \bibinfo{author}{Shlens, J.},
  \bibinfo{author}{Szegedy, C.}, \bibinfo{year}{2015}.
\newblock \bibinfo{title}{Explaining and harnessing adversarial examples}, in:
  \bibinfo{booktitle}{ICLR}.
%Type = Inproceedings
\bibitem[{Goyal et~al.(2020)Goyal, Raghunathan, Jain, Simhadri and
  Jain}]{Goyal2020DROCCDR}
\bibinfo{author}{Goyal, S.}, \bibinfo{author}{Raghunathan, A.},
  \bibinfo{author}{Jain, M.}, \bibinfo{author}{Simhadri, H.V.},
  \bibinfo{author}{Jain, P.}, \bibinfo{year}{2020}.
\newblock \bibinfo{title}{{DROCC:} deep robust one-class classification}, in:
  \bibinfo{booktitle}{ICML}.
%Type = Article
\bibitem[{Hallaji et~al.(2023)Hallaji, Razavi-Far, Saif and
  Herrera-Viedma}]{HALLAJI2023110384}
\bibinfo{author}{Hallaji, E.}, \bibinfo{author}{Razavi-Far, R.},
  \bibinfo{author}{Saif, M.}, \bibinfo{author}{Herrera-Viedma, E.},
  \bibinfo{year}{2023}.
\newblock \bibinfo{title}{Label noise analysis meets adversarial training: A
  defense against label poisoning in federated learning}.
\newblock \bibinfo{journal}{Knowledge-Based Systems} .
%Type = Inproceedings
\bibitem[{He et~al.(2016)He, Zhang, Ren and Sun}]{He2016DeepRL}
\bibinfo{author}{He, K.}, \bibinfo{author}{Zhang, X.}, \bibinfo{author}{Ren,
  S.}, \bibinfo{author}{Sun, J.}, \bibinfo{year}{2016}.
\newblock \bibinfo{title}{Deep residual learning for image recognition}, in:
  \bibinfo{booktitle}{CVPR}.
%Type = Article
\bibitem[{Hospedales et~al.(2021)Hospedales, Antoniou, Micaelli and
  Storkey}]{hospedales2021meta}
\bibinfo{author}{Hospedales, T.}, \bibinfo{author}{Antoniou, A.},
  \bibinfo{author}{Micaelli, P.}, \bibinfo{author}{Storkey, A.},
  \bibinfo{year}{2021}.
\newblock \bibinfo{title}{Meta-learning in neural networks: A survey}.
\newblock \bibinfo{journal}{IEEE transactions on pattern analysis and machine
  intelligence} .
%Type = Article
\bibitem[{Jaiswal et~al.(2020)Jaiswal, Babu, Zadeh, Banerjee and
  Makedon}]{jaiswal2020survey}
\bibinfo{author}{Jaiswal, A.}, \bibinfo{author}{Babu, A.R.},
  \bibinfo{author}{Zadeh, M.Z.}, \bibinfo{author}{Banerjee, D.},
  \bibinfo{author}{Makedon, F.}, \bibinfo{year}{2020}.
\newblock \bibinfo{title}{A survey on contrastive self-supervised learning}.
\newblock \bibinfo{journal}{Technologies} .
%Type = Article
\bibitem[{Kadir and Brady(2001)}]{Kadir2004SaliencySA}
\bibinfo{author}{Kadir, T.}, \bibinfo{author}{Brady, M.}, \bibinfo{year}{2001}.
\newblock \bibinfo{title}{Saliency, scale and image description}.
\newblock \bibinfo{journal}{Int. J. Comput. Vis.} .
%Type = Article
\bibitem[{Kannan et~al.(2018)Kannan, Kurakin and
  Goodfellow}]{Kannan2018AdversarialLP}
\bibinfo{author}{Kannan, H.}, \bibinfo{author}{Kurakin, A.},
  \bibinfo{author}{Goodfellow, I.J.}, \bibinfo{year}{2018}.
\newblock \bibinfo{title}{Adversarial logit pairing}.
\newblock \bibinfo{journal}{CoRR} .
%Type = Article
\bibitem[{Ke et~al.(2024)Ke, Zheng, Li, He, Li and Min}]{KE2024111909}
\bibinfo{author}{Ke, W.}, \bibinfo{author}{Zheng, D.}, \bibinfo{author}{Li,
  X.}, \bibinfo{author}{He, Y.}, \bibinfo{author}{Li, T.},
  \bibinfo{author}{Min, F.}, \bibinfo{year}{2024}.
\newblock \bibinfo{title}{Improving the transferability of adversarial examples
  through neighborhood attribution}.
\newblock \bibinfo{journal}{Knowledge-Based Systems} .
%Type = Inproceedings
\bibitem[{Kingma and Ba(2015)}]{DBLP:journals/corr/KingmaB14}
\bibinfo{author}{Kingma, D.P.}, \bibinfo{author}{Ba, J.}, \bibinfo{year}{2015}.
\newblock \bibinfo{title}{Adam: {A} method for stochastic optimization}, in:
  \bibinfo{booktitle}{ICLR}.
%Type = Article
\bibitem[{Krizhevsky et~al.(2009)Krizhevsky, Hinton
  et~al.}]{Krizhevsky2009LearningML}
\bibinfo{author}{Krizhevsky, A.}, \bibinfo{author}{Hinton, G.}, et~al.,
  \bibinfo{year}{2009}.
\newblock \bibinfo{title}{Learning multiple layers of features from tiny
  images} .
%Type = Inproceedings
\bibitem[{Kurakin et~al.(2017)Kurakin, Goodfellow and
  Bengio}]{kurakin2016adversarial}
\bibinfo{author}{Kurakin, A.}, \bibinfo{author}{Goodfellow, I.J.},
  \bibinfo{author}{Bengio, S.}, \bibinfo{year}{2017}.
\newblock \bibinfo{title}{Adversarial examples in the physical world}, in:
  \bibinfo{booktitle}{ICLR}.
%Type = Inproceedings
\bibitem[{Li et~al.(2022)Li, Zhu, Jia, Jiang, Xia and Cao}]{li2022defending}
\bibinfo{author}{Li, Y.}, \bibinfo{author}{Zhu, L.}, \bibinfo{author}{Jia, X.},
  \bibinfo{author}{Jiang, Y.}, \bibinfo{author}{Xia, S.T.},
  \bibinfo{author}{Cao, X.}, \bibinfo{year}{2022}.
\newblock \bibinfo{title}{Defending against model stealing via verifying
  embedded external features}, in: \bibinfo{booktitle}{AAAI}.
%Type = Article
\bibitem[{Liu et~al.(2021)Liu, Zhang, Hou, Mian, Wang, Zhang and
  Tang}]{liu2021self}
\bibinfo{author}{Liu, X.}, \bibinfo{author}{Zhang, F.}, \bibinfo{author}{Hou,
  Z.}, \bibinfo{author}{Mian, L.}, \bibinfo{author}{Wang, Z.},
  \bibinfo{author}{Zhang, J.}, \bibinfo{author}{Tang, J.},
  \bibinfo{year}{2021}.
\newblock \bibinfo{title}{Self-supervised learning: Generative or contrastive}.
\newblock \bibinfo{journal}{IEEE transactions on knowledge and data
  engineering} .
%Type = Article
\bibitem[{Van~der Maaten and Hinton(2008)}]{Maaten2008VisualizingDU}
\bibinfo{author}{Van~der Maaten, L.}, \bibinfo{author}{Hinton, G.},
  \bibinfo{year}{2008}.
\newblock \bibinfo{title}{Visualizing data using t-sne}.
\newblock \bibinfo{journal}{Journal of Machine Learning Research} .
%Type = Article
\bibitem[{McInnes et~al.(2018)McInnes, Healy, Saul and
  Gro{\ss}berger}]{DBLP:journals/jossw/McInnesHSG18}
\bibinfo{author}{McInnes, L.}, \bibinfo{author}{Healy, J.},
  \bibinfo{author}{Saul, N.}, \bibinfo{author}{Gro{\ss}berger, L.},
  \bibinfo{year}{2018}.
\newblock \bibinfo{title}{{UMAP:} uniform manifold approximation and
  projection}.
\newblock \bibinfo{journal}{J. Open Source Softw.} .
%Type = Inproceedings
\bibitem[{Miko{\l}ajczyk and Grochowski(2018)}]{Mikoajczyk2018DataAF}
\bibinfo{author}{Miko{\l}ajczyk, A.}, \bibinfo{author}{Grochowski, M.},
  \bibinfo{year}{2018}.
\newblock \bibinfo{title}{Data augmentation for improving deep learning in
  image classification problem}, in: \bibinfo{booktitle}{international
  interdisciplinary PhD workshop (IIPhDW)}.
%Type = Article
\bibitem[{Noack et~al.(2019)Noack, Ahern, Dou and Li}]{Noack2019DoesIO}
\bibinfo{author}{Noack, A.}, \bibinfo{author}{Ahern, I.}, \bibinfo{author}{Dou,
  D.}, \bibinfo{author}{Li, B.}, \bibinfo{year}{2019}.
\newblock \bibinfo{title}{Does interpretability of neural networks imply
  adversarial robustness?}
\newblock \bibinfo{journal}{CoRR} .
%Type = Article
\bibitem[{Qian et~al.(2022)Qian, Huang, Wang and Zhang}]{qian2022survey}
\bibinfo{author}{Qian, Z.}, \bibinfo{author}{Huang, K.}, \bibinfo{author}{Wang,
  Q.F.}, \bibinfo{author}{Zhang, X.Y.}, \bibinfo{year}{2022}.
\newblock \bibinfo{title}{A survey of robust adversarial training in pattern
  recognition: Fundamental, theory, and methodologies}.
\newblock \bibinfo{journal}{Pattern Recognition} .
%Type = Inproceedings
\bibitem[{Ribeiro et~al.(2016)Ribeiro, Singh and Guestrin}]{Ribeiro2016WhySI}
\bibinfo{author}{Ribeiro, M.T.}, \bibinfo{author}{Singh, S.},
  \bibinfo{author}{Guestrin, C.}, \bibinfo{year}{2016}.
\newblock \bibinfo{title}{"why should {I} trust you?": Explaining the
  predictions of any classifier}, in: \bibinfo{booktitle}{SIGKDD}.
%Type = Article
\bibitem[{Russakovsky et~al.(2015)Russakovsky, Deng, Su, Krause, Satheesh, Ma,
  Huang, Karpathy, Khosla, Bernstein, Berg and
  Fei{-}Fei}]{Russakovsky2015ImageNetLS}
\bibinfo{author}{Russakovsky, O.}, \bibinfo{author}{Deng, J.},
  \bibinfo{author}{Su, H.}, \bibinfo{author}{Krause, J.},
  \bibinfo{author}{Satheesh, S.}, \bibinfo{author}{Ma, S.},
  \bibinfo{author}{Huang, Z.}, \bibinfo{author}{Karpathy, A.},
  \bibinfo{author}{Khosla, A.}, \bibinfo{author}{Bernstein, M.S.},
  \bibinfo{author}{Berg, A.C.}, \bibinfo{author}{Fei{-}Fei, L.},
  \bibinfo{year}{2015}.
\newblock \bibinfo{title}{Imagenet large scale visual recognition challenge}.
\newblock \bibinfo{journal}{Int. J. Comput. Vis.} .
%Type = Article
\bibitem[{Salehi et~al.(2021)Salehi, Arya, Pajoum, Otoofi, Shaeiri, Rohban and
  Rabiee}]{Salehi2020ARAEAR}
\bibinfo{author}{Salehi, M.}, \bibinfo{author}{Arya, A.},
  \bibinfo{author}{Pajoum, B.}, \bibinfo{author}{Otoofi, M.},
  \bibinfo{author}{Shaeiri, A.}, \bibinfo{author}{Rohban, M.H.},
  \bibinfo{author}{Rabiee, H.R.}, \bibinfo{year}{2021}.
\newblock \bibinfo{title}{{ARAE:} adversarially robust training of autoencoders
  improves novelty detection}.
\newblock \bibinfo{journal}{Neural Networks} .
%Type = Inproceedings
\bibitem[{Salman et~al.(2020)Salman, Ilyas, Engstrom, Kapoor and
  Madry}]{Salman2020DoAR}
\bibinfo{author}{Salman, H.}, \bibinfo{author}{Ilyas, A.},
  \bibinfo{author}{Engstrom, L.}, \bibinfo{author}{Kapoor, A.},
  \bibinfo{author}{Madry, A.}, \bibinfo{year}{2020}.
\newblock \bibinfo{title}{Do adversarially robust imagenet models transfer
  better?}, in: \bibinfo{booktitle}{NeurIPS}.
%Type = Article
\bibitem[{Santurkar et~al.(2019)Santurkar, Tsipras, Tran, Ilyas, Engstrom and
  Madry}]{Santurkar2019ComputerVW}
\bibinfo{author}{Santurkar, S.}, \bibinfo{author}{Tsipras, D.},
  \bibinfo{author}{Tran, B.}, \bibinfo{author}{Ilyas, A.},
  \bibinfo{author}{Engstrom, L.}, \bibinfo{author}{Madry, A.},
  \bibinfo{year}{2019}.
\newblock \bibinfo{title}{Computer vision with a single (robust) classifier}.
\newblock \bibinfo{journal}{CoRR} .
%Type = Inproceedings
\bibitem[{Schroff et~al.(2015)Schroff, Kalenichenko and
  Philbin}]{schroff2015facenet}
\bibinfo{author}{Schroff, F.}, \bibinfo{author}{Kalenichenko, D.},
  \bibinfo{author}{Philbin, J.}, \bibinfo{year}{2015}.
\newblock \bibinfo{title}{Facenet: A unified embedding for face recognition and
  clustering}, in: \bibinfo{booktitle}{CVPR}.
%Type = Article
\bibitem[{Selvaraju et~al.(2020)Selvaraju, Cogswell, Das, Vedantam, Parikh and
  Batra}]{Selvaraju2019GradCAMVE}
\bibinfo{author}{Selvaraju, R.R.}, \bibinfo{author}{Cogswell, M.},
  \bibinfo{author}{Das, A.}, \bibinfo{author}{Vedantam, R.},
  \bibinfo{author}{Parikh, D.}, \bibinfo{author}{Batra, D.},
  \bibinfo{year}{2020}.
\newblock \bibinfo{title}{Grad-cam: Visual explanations from deep networks via
  gradient-based localization}.
\newblock \bibinfo{journal}{Int. J. Comput. Vis.} .
%Type = Article
\bibitem[{Shorten and Khoshgoftaar(2019)}]{DBLP:journals/jbd/ShortenK19}
\bibinfo{author}{Shorten, C.}, \bibinfo{author}{Khoshgoftaar, T.M.},
  \bibinfo{year}{2019}.
\newblock \bibinfo{title}{A survey on image data augmentation for deep
  learning}.
\newblock \bibinfo{journal}{J. Big Data} .
%Type = Article
\bibitem[{Sundararajan et~al.(2016)Sundararajan, Taly and
  Yan}]{DBLP:journals/corr/SundararajanTY16}
\bibinfo{author}{Sundararajan, M.}, \bibinfo{author}{Taly, A.},
  \bibinfo{author}{Yan, Q.}, \bibinfo{year}{2016}.
\newblock \bibinfo{title}{Gradients of counterfactuals}.
\newblock \bibinfo{journal}{CoRR} .
%Type = Inproceedings
\bibitem[{Terzi et~al.(2021)Terzi, Achille, Maggipinto and
  Susto}]{Terzi2020AdversarialTR}
\bibinfo{author}{Terzi, M.}, \bibinfo{author}{Achille, A.},
  \bibinfo{author}{Maggipinto, M.}, \bibinfo{author}{Susto, G.A.},
  \bibinfo{year}{2021}.
\newblock \bibinfo{title}{Adversarial training reduces information and improves
  transferability}, in: \bibinfo{booktitle}{AAAI}.
%Type = Inproceedings
\bibitem[{Utrera et~al.(2021)Utrera, Kravitz, Erichson, Khanna and
  Mahoney}]{Utrera2020AdversariallyTrainedDN}
\bibinfo{author}{Utrera, F.}, \bibinfo{author}{Kravitz, E.},
  \bibinfo{author}{Erichson, N.B.}, \bibinfo{author}{Khanna, R.},
  \bibinfo{author}{Mahoney, M.W.}, \bibinfo{year}{2021}.
\newblock \bibinfo{title}{Adversarially-trained deep nets transfer better:
  Illustration on image classification}, in: \bibinfo{booktitle}{ICLR}.
%Type = Article
\bibitem[{Verma et~al.(2020)Verma, Dickerson and
  Hines}]{verma2020counterfactual}
\bibinfo{author}{Verma, S.}, \bibinfo{author}{Dickerson, J.},
  \bibinfo{author}{Hines, K.}, \bibinfo{year}{2020}.
\newblock \bibinfo{title}{Counterfactual explanations for machine learning: A
  review}.
\newblock \bibinfo{journal}{CoRR} .
%Type = Article
\bibitem[{Wang et~al.(2021)Wang, Chen, Tang, Yue, Zhu, Zeng and
  Wang}]{WANG2021107141}
\bibinfo{author}{Wang, L.}, \bibinfo{author}{Chen, X.}, \bibinfo{author}{Tang,
  R.}, \bibinfo{author}{Yue, Y.}, \bibinfo{author}{Zhu, Y.},
  \bibinfo{author}{Zeng, X.}, \bibinfo{author}{Wang, W.}, \bibinfo{year}{2021}.
\newblock \bibinfo{title}{Improving adversarial robustness of deep neural
  networks by using semantic information}.
\newblock \bibinfo{journal}{Knowledge-Based Systems} .
%Type = Article
\bibitem[{Wang et~al.(2023)Wang, Zhao, Lu, Wang and
  Zhang}]{DBLP:journals/isci/WangZLWZ23}
\bibinfo{author}{Wang, L.}, \bibinfo{author}{Zhao, X.}, \bibinfo{author}{Lu,
  Z.}, \bibinfo{author}{Wang, L.}, \bibinfo{author}{Zhang, S.},
  \bibinfo{year}{2023}.
\newblock \bibinfo{title}{Enhancing privacy preservation and trustworthiness
  for decentralized federated learning}.
\newblock \bibinfo{journal}{Information Sciences} .
%Type = Article
\bibitem[{Wang et~al.(2022)Wang, Wang, Liang, Zhao, Huang, Xu, Dai and
  Miao}]{wang2022deep}
\bibinfo{author}{Wang, X.}, \bibinfo{author}{Wang, S.}, \bibinfo{author}{Liang,
  X.}, \bibinfo{author}{Zhao, D.}, \bibinfo{author}{Huang, J.},
  \bibinfo{author}{Xu, X.}, \bibinfo{author}{Dai, B.}, \bibinfo{author}{Miao,
  Q.}, \bibinfo{year}{2022}.
\newblock \bibinfo{title}{Deep reinforcement learning: A survey}.
\newblock \bibinfo{journal}{IEEE Transactions on Neural Networks and Learning
  Systems} .
%Type = Article
\bibitem[{Yi et~al.(2022)Yi, Chen, Xu, Liu, Jiang and Tan}]{YI2022108831}
\bibinfo{author}{Yi, C.}, \bibinfo{author}{Chen, H.}, \bibinfo{author}{Xu, Y.},
  \bibinfo{author}{Liu, Y.}, \bibinfo{author}{Jiang, L.}, \bibinfo{author}{Tan,
  H.}, \bibinfo{year}{2022}.
\newblock \bibinfo{title}{Atpl: Mutually enhanced adversarial training and
  pseudo labeling for unsupervised domain adaptation}.
\newblock \bibinfo{journal}{Knowledge-Based Systems} .
%Type = Article
\bibitem[{Zhang et~al.(2023)Zhang, Zhu, Li, Zhou and Yu}]{ZHANG2023110777}
\bibinfo{author}{Zhang, T.}, \bibinfo{author}{Zhu, T.}, \bibinfo{author}{Li,
  J.}, \bibinfo{author}{Zhou, W.}, \bibinfo{author}{Yu, P.S.},
  \bibinfo{year}{2023}.
\newblock \bibinfo{title}{Revisiting model fairness via adversarial examples}.
\newblock \bibinfo{journal}{Knowledge-Based Systems} .
%Type = Inproceedings
\bibitem[{Zhang and Zhu(2019)}]{Zhang2019InterpretingAT}
\bibinfo{author}{Zhang, T.}, \bibinfo{author}{Zhu, Z.}, \bibinfo{year}{2019}.
\newblock \bibinfo{title}{Interpreting adversarially trained convolutional
  neural networks}, in: \bibinfo{booktitle}{ICML}.
%Type = Inproceedings
\bibitem[{Zheng et~al.(2016)Zheng, Song, Leung and
  Goodfellow}]{Zheng2016ImprovingTR}
\bibinfo{author}{Zheng, S.}, \bibinfo{author}{Song, Y.},
  \bibinfo{author}{Leung, T.}, \bibinfo{author}{Goodfellow, I.J.},
  \bibinfo{year}{2016}.
\newblock \bibinfo{title}{Improving the robustness of deep neural networks via
  stability training}, in: \bibinfo{booktitle}{CVPR}.

\end{thebibliography}

\end{document}